\documentclass[11pt]{article}

\usepackage[final]{acl}

\usepackage{times} 
\usepackage{latexsym}
\usepackage{comment}
\usepackage{amsmath}
\usepackage{amssymb}
\usepackage{graphicx}
\usepackage{booktabs}
\usepackage{appendix}
\usepackage{chngcntr}
\usepackage{float} 
\usepackage{cuted} 
\usepackage[table]{xcolor}
\usepackage{multirow}
\usepackage{booktabs}
\usepackage{subcaption}
\usepackage{color}
\usepackage{bbm}
\usepackage[table]{xcolor}

\usepackage[T1]{fontenc}

\usepackage[utf8]{inputenc}

\usepackage{microtype}

\usepackage{inconsolata}
\usepackage{xcolor}
\usepackage[table]{xcolor}

\newcommand{\rebuttal}[1]{\textcolor{black}{#1}}
\newcommand{\camera}[1]{\textcolor{black}{#1}}
\newcommand{\finaledits}[1]{\textcolor{black}{#1}}
\definecolor{dkgreen}{rgb}{0,0.6,0}
\definecolor{gray}{rgb}{0.5,0.5,0.5}
\definecolor{mauve}{rgb}{0.58,0,0.82}
\definecolor{lightblue}{rgb}{0.9, 0.95, 1.0}
\definecolor{lightgray}{rgb}{0.9, 0.9, 0.9}
\definecolor{softblue}{RGB}{70,130,180} 
\definecolor{color5L}{HTML}{00008B}
\definecolor{color4L}{HTML}{2E8B57}
\definecolor{color3L}{HTML}{E68A00}
\definecolor{color2L}{HTML}{7f5539}
\definecolor{color1L}{HTML}{B22222}
\definecolor{-}{rgb}{0.70,0.13,0.13}
\definecolor{+}{rgb}{0.0, 0.6, 0.3} 
\newcommand{\pct}[1]{\textcolor{softblue}{(#1)}}

\title{Identifying Crucial Attention Heads for Multilingual Language Models: Retrieval and Retrieval-Transition Heads}

\author{
\textbf{Shaswat Patel}\thanks{Equal contribution.}\thanks{Corresponding author: spp9399@nyu.edu and vt2369@nyu.edu} \quad
\textbf{Vishvesh Trivedi}\footnotemark[1]\footnotemark[2] \quad
\textbf{Yue Han} \quad
\textbf{Yihuai Hong} \quad
\textbf{Eunsol Choi} \\
New York University
}
  

\begin{document}
\maketitle

\begin{abstract}

Retrieval heads, a subset of attention heads in Transformers, were studied in English, showing its crucial role in retrieving information from the context. We expand the study of retrieval heads to multilingual context and find that while nearly half of all retrieval heads are often shared across multiple languages, language-specific retrieval heads also emerge. We further design a cross-lingual needle-in-the-haystack task, to identify \textit{Retrieval Transition Heads (RTH)} that retrieve key information and further map it to target-language output. Our experiments reveal that RTHs do not always overlap with RH, and can be more vital for Chain-of-Thought reasoning in multilingual LLMs. Across four multilingual benchmarks (MMLU-ProX, MGSM, MLQA, and XQuAD) and two model families (Qwen2.5-7B Instruct and Llama3.1-8B Instruct), we demonstrate that masking RTH induces bigger performance drop than masking Retrieval Heads (RH). Specifically, for Llama3.1-8B Instruct, masking the top-25 RTHs results in an average 37.3-point drop (against 19.5-point drop from RH masking) in reasoning accuracy (MMLU-ProX, MGSM) and 9.0 F1-score reduction (against 5.1 drop for RH) in extractive QA (MLQA, XQuAD). Our work advances the understanding of multilingual LMs by isolating the attention heads responsible for mapping to target languages. Code available at \url{https://github.com/NerdyVisky/multilingual-retrieval-transition-heads}


\end{abstract}

\section{Introduction}

As LLMs are deployed globally, their ability to retrieve, reason, and respond across multiple languages has become crucial. Recent research has linked specific behaviors to internal circuits \cite{zhou2024pre, ferrando2024know}, identifying \textit{retrieval heads} (RH) as the sparse subset of attention heads responsible for extracting facts from long contexts \cite{wu2024retrieval}. In this work, we study such retrieval heads in multilingual settings.

Evaluating three model families on five languages (English, German, Chinese, Swahili and Afrikaans), we identify both language-shared and language-exclusive retrieval heads. We also report different proportions of language-sharing behavior of retrieval heads for different model families. Qwen2.5-7B Instruct \cite{qwen2025qwen25technicalreport} exhibits 57.6\% overlap across all five languages with only 13.3\% language-specific heads, while retrieval heads in Phi3.5-3B MiniInstruct \cite{abdin2024phi3technicalreporthighly} show only 41.5\% overlap among five languages with significantly higher language-specific heads at 25.9\% (\S{\ref{sec:Multilingual_RH}}). We observe a rough correlation between the number of non-English pretraining tokens and the degree of language-sharing nature of its retrieval heads i.e. more multilingual training corresponds to more language-shared behavior, similar to analogous findings for MLP/neurons  \cite{chen2025emergence}.


\rebuttal{Multilingual models are observed to reason using English-centric latent space, independent of the language used for generation \cite{wendler2024Dollama,wang2025lost,dumas-etal-2025-separating}.} \finaledits{Building on this observation, we introduce a cross-lingual needle-in-the-haystack setting: the evidence context in a high-resource language and the query in the target language.} 
We define these  Retrieval-Transition Heads (RTH), by measuring whether a head’s attention aligns target-language output tokens with their corresponding source-language "needle" counterparts (\S{\ref{sec:RTH_Discovery}}). \rebuttal{We observe that RTHs can be more crucial for multilingual reasoning than RHs by ablating them. Masking a sparse set of RTHs triggers a catastrophic performance drop—most notably a 54-point drop in MGSM for Llama-3.1 averaged across four languages, about 2.5$\times$ the drop seen in standard RH masking (\S{\ref{sec:RTH_experiments}})}. Qualitatively, RTH masking decreases language coherence when compared to RH masking (Table~\ref{tab:failure_mode_counts_with_pct_rh_rth}). 
Our contributions are as follows:
\begin{itemize}
    \item We analyze retrieval heads in multilingual LLMs, finding that they decompose into language-shared and language-exclusive subsets, with significant variance across model families.
    \item We define the Retrieval-Transition Score to measure cross-lingual correspondence at the head level and characterize the Retrieval-Transition Heads responsible for mapping retrieved evidence to target-language synthesis.
    \item Through masking experiments on several multilingual reasoning benchmarks, we identify RTH is more crucial than RH for achieving high performance. 
\end{itemize}


\section{Related Work}
\textbf{Role of attention heads} Various works studied Multi-Head Attention (MHA) in Transformer architectures and discovered distinct functions performed by a certain set of attention heads \cite{zheng2024attention,gould2024successor}. 
For example, \citet{mcdougall2024copy} studies copy suppression, the functionality of the attention head to reduce the prediction of a token that has appeared previously in the context. 
\citet{olsson2022context} discovered \textit{Induction heads}, which implement a "match and copy" mechanism to complete patterns found in the preceding context. 

Most relevant to our work, \citet{wu2024retrieval} identify \textit{Retrieval Heads} as the primary drivers of in-context learning and retrieval. They note that masking the 50 most influential retrieval heads leads up to 50\% drop in performance tasks requiring Chain-of-Thought reasoning. This contributed to subsequent applications of the insight, such as head-level KV cache compression \cite{fu2025not} and improving contextual faithfulness in long-horizon reasoning \cite{huang2025improving}.

\paragraph{Understanding Multilingual Language Models} 
Despite an English-centric bias in training data, MLMs demonstrate surprisingly good multilingual proficiency \cite{zhao2025babel}. \citet{wendler2024Dollama} investigated this via logit lens and causal interventions. They show that while the initial and final layers operate in the input/output language, the middle layers transition into an English-centric latent concept space. 
Following this study, \citet{wang2025lost} explored cross-lingual inconsistencies, where models successfully retrieve parametric facts when prompted in English but fail in other languages. They attribute failure in mapping language-agnostic concepts back into the specific non-English token space. Recent work by \citet{chen2025emergence} reports that as models scale and are trained more multi-lingually, they move away from English-centric behavior and develop a core language-agnostic parameter space composed of MLP heads termed as `shared neurons'. Further, \citet{zhang2025the} reported that MLMs share circuits for identical syntactic processes while employing distinct attention heads and feed-forward layers for language-specific linguistic processes. We investigate whether retrieval heads are similarly partitioned and compare how this attribute varies with model families.


\paragraph{Attention heads in multilingual LLMs} Recent studies have localized language-specific heads and demonstrated how causal interventions can shift a model's language focus \cite{liu2025focusing} or trigger direct translation \cite{liu2026token}. While we share the goal of identifying heads responsible for multilingual performance, prior efforts focus largely on translation or short-context QA. These approaches offer limited insight into modern workflows requiring sustained coherence across long Chain-of-Thought (CoT) sequences. We introduce Retrieval-Transition Heads (RTH), which are analogous to retrieval heads \cite{wu2024retrieval} but specialized for multilingual contexts such that masking them causes failures in linguistic coherence and context grounding even when the model attempts to answer in the correct target language.

\section{Retrieval Heads in Diverse Languages} 
\label{sec:Multilingual_RH}
\subsection{Background}
We describe the process for identifying retrieval heads from \citet{wu2024retrieval} below. 
\textbf{\\ Needle-In-A-Haystack:} NIAH task consists of a tuple $(c, q, k)$ where $c$ is the long-context text (a.k.a. haystack), $q$ is the question and $k$ is the answer related to the question (a.k.a. needle). Generally, $q$ is framed so that the model has no parametric knowledge and has to retrieve $k$ to answer the question. 

\noindent \textbf{Retrieval score (RS): } Each attention head's retrieval score is computed as the proportion of decoded tokens part of the needle sentence where the head assigned the most attention weight. More formally, the retrieval score for an attention head per example $RS_h$ as:


\[
RS_h = \frac{1}{N}\sum_{t=1}^{M} \mathcal{N}^t,
\]
\[
\text{where,} \quad 
\mathcal{N}^t =
 \mathbbm{1}\left( \arg\max_{i} a_{h,i}^t = \mathcal{O} \right)
\]

\noindent Here, $RS_h$ is the retrieval score for attention head $h$, $N$ is the number of tokens in the needle sentence, $M$ is the max decoding limit, set to $M=50$,  $\mathbbm{1}(\cdot)$ is the indicator function, $a_{h,i}^{(t)}$ is the attention weight assigned to input token, part of the context, $i$ by head $h$ at decoding timestep $t$, $L$ is the total length of the input context, and $\mathcal{O}$ is the last decoded output token. For example, given a needle sentence of length $n$ tokens, when decoding the word "Francisco", if a head assigns the most attention weight to the word "Francisco" \textit{within the needle}, then the score increments by $\frac{1}{n}$ for that head. \camera{Following the same definition as \citet{wu2024retrieval} we consider all heads with $RS_h \ge 0.01$ as a Retrieval head.}

\subsection{Multilingual Experimental Setting} \label{sec:3_models_and_langs}


\begin{figure}
    \centering
    \includegraphics[width=0.85\linewidth]{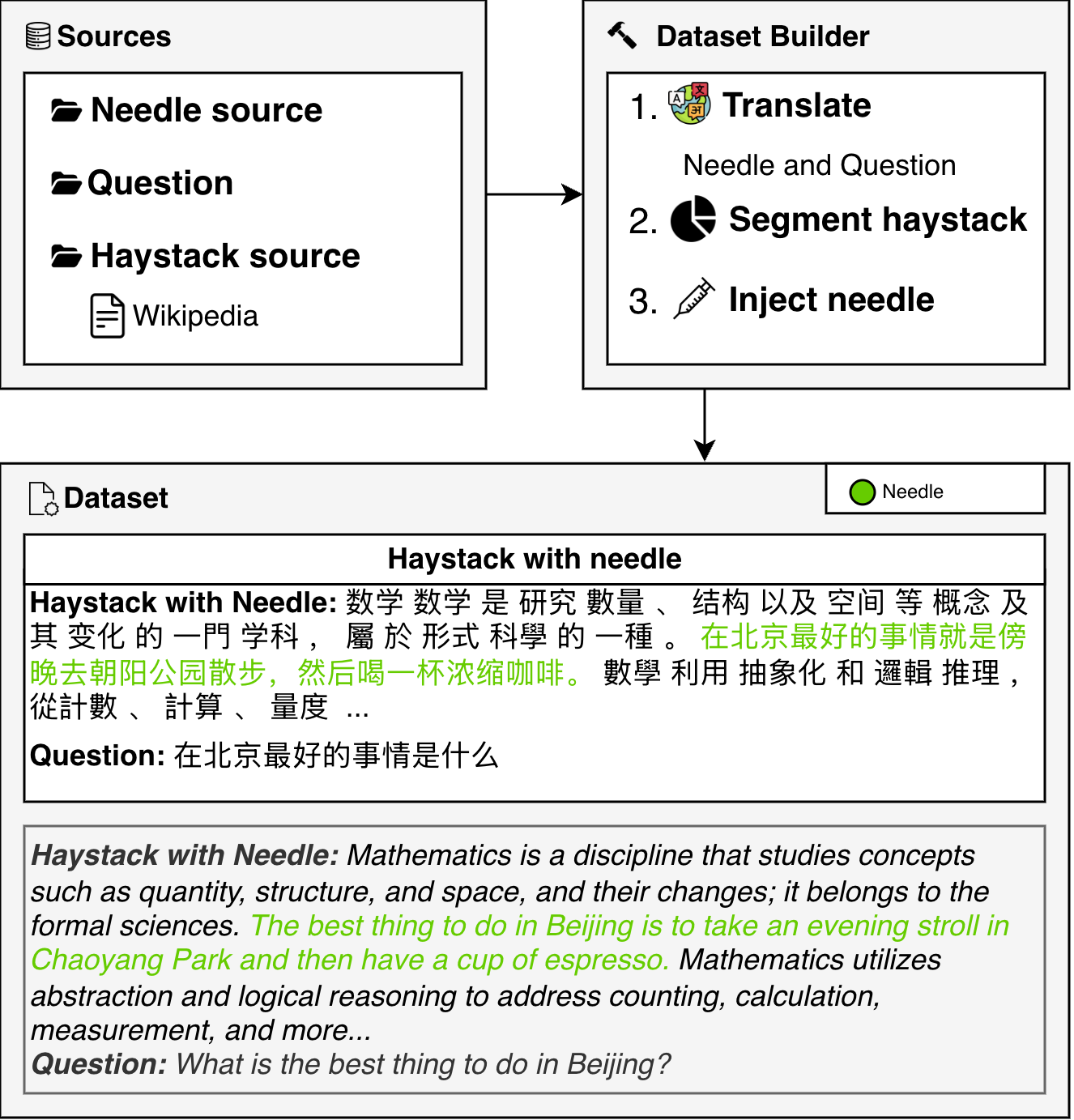}
    \caption{Multilingual NIAH dataset construction pipeline. We sample needles, questions and haystacks, translate the needle and question into the target language, segment the haystack, and inject the translated needle to produce the final evaluation instances.}
    \label{fig:niah_data_pipeline}
\end{figure}

\begin{figure*}[t!]
\centering
\begin{minipage}{0.48\linewidth}
    \centering
    \includegraphics[width=\linewidth]{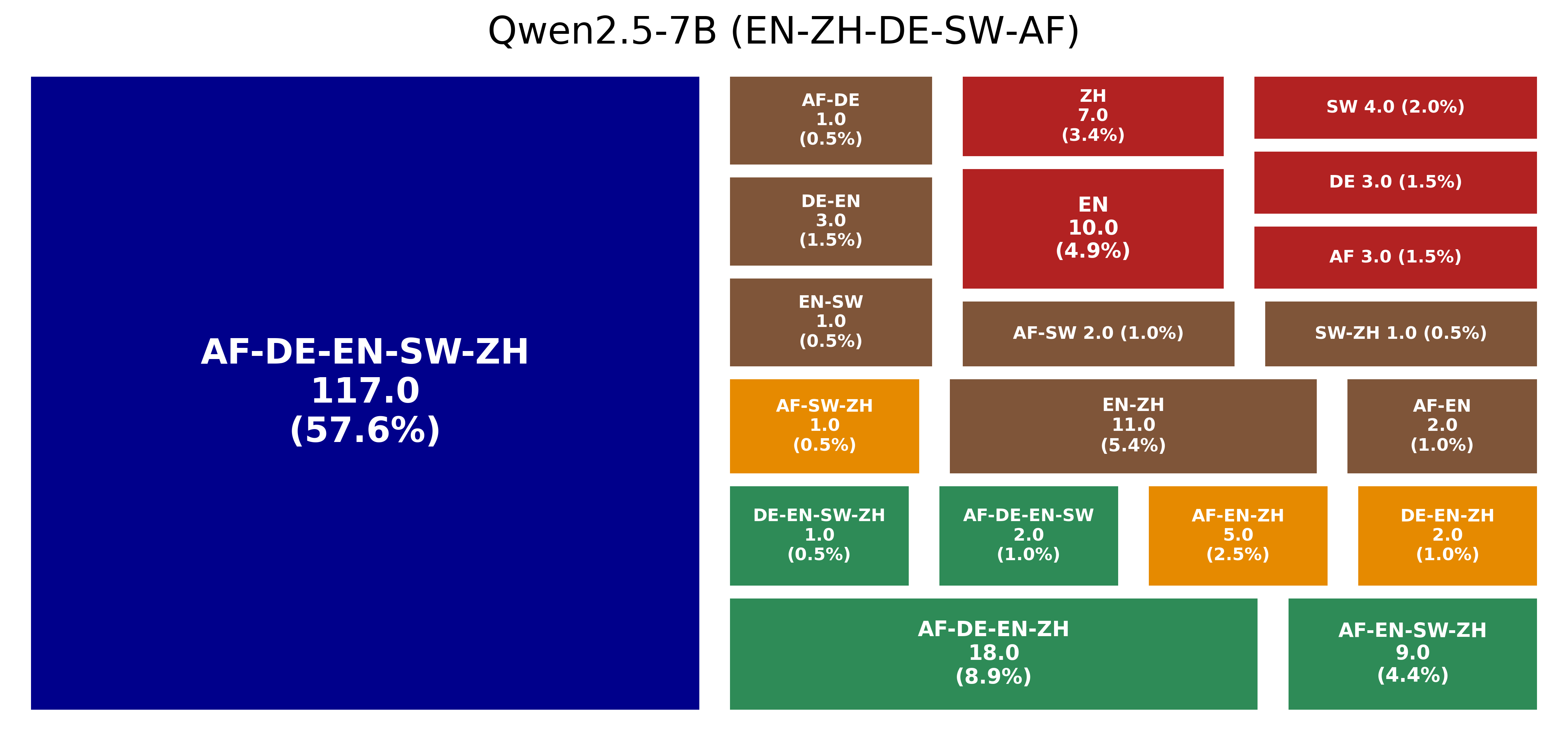}
\end{minipage}
\hfill
\begin{minipage}{0.48\linewidth}
    \centering
    \includegraphics[width=\linewidth]{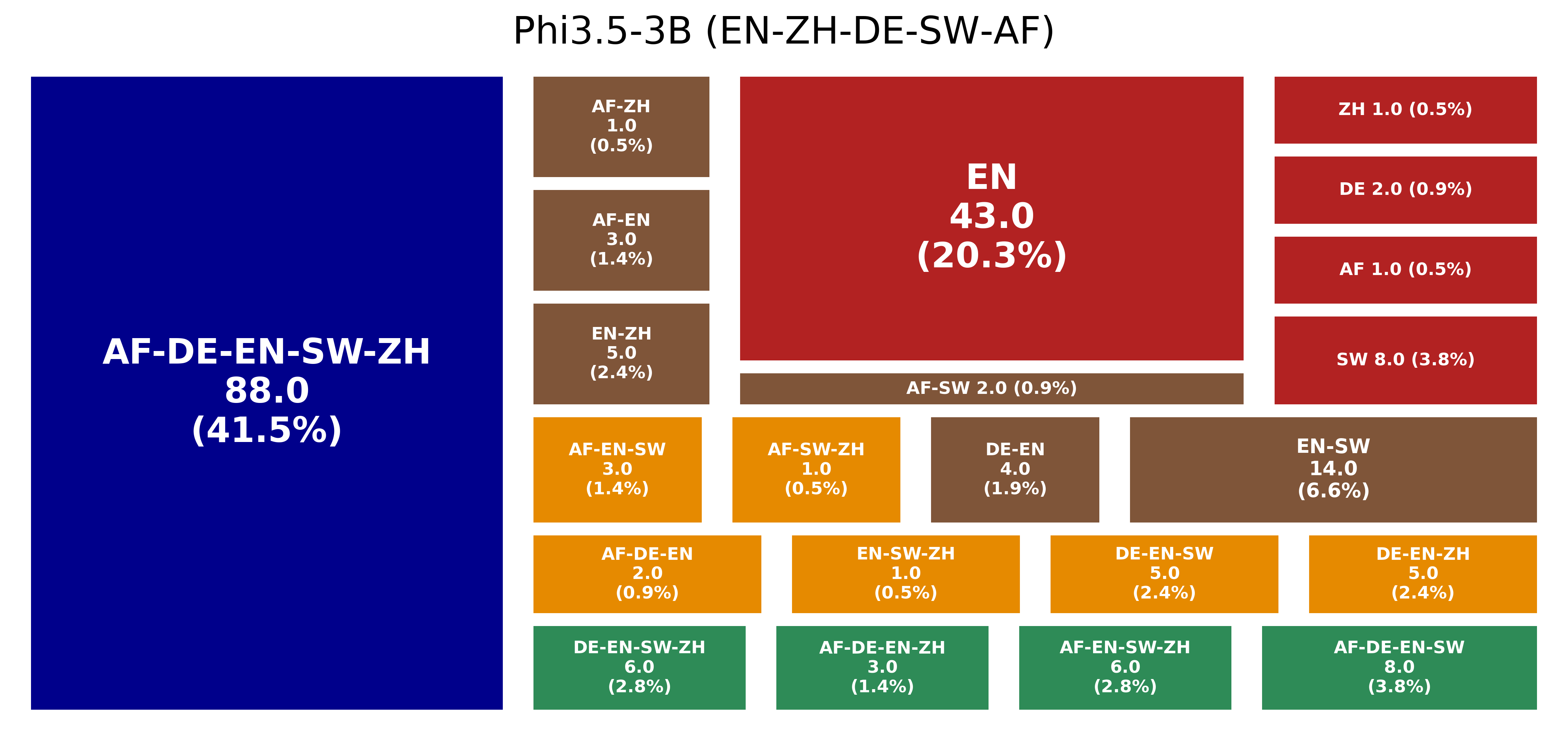}
\end{minipage}

\vspace{1.0em} 

\begin{minipage}{0.48\linewidth}
    \centering
    \includegraphics[width=\linewidth]{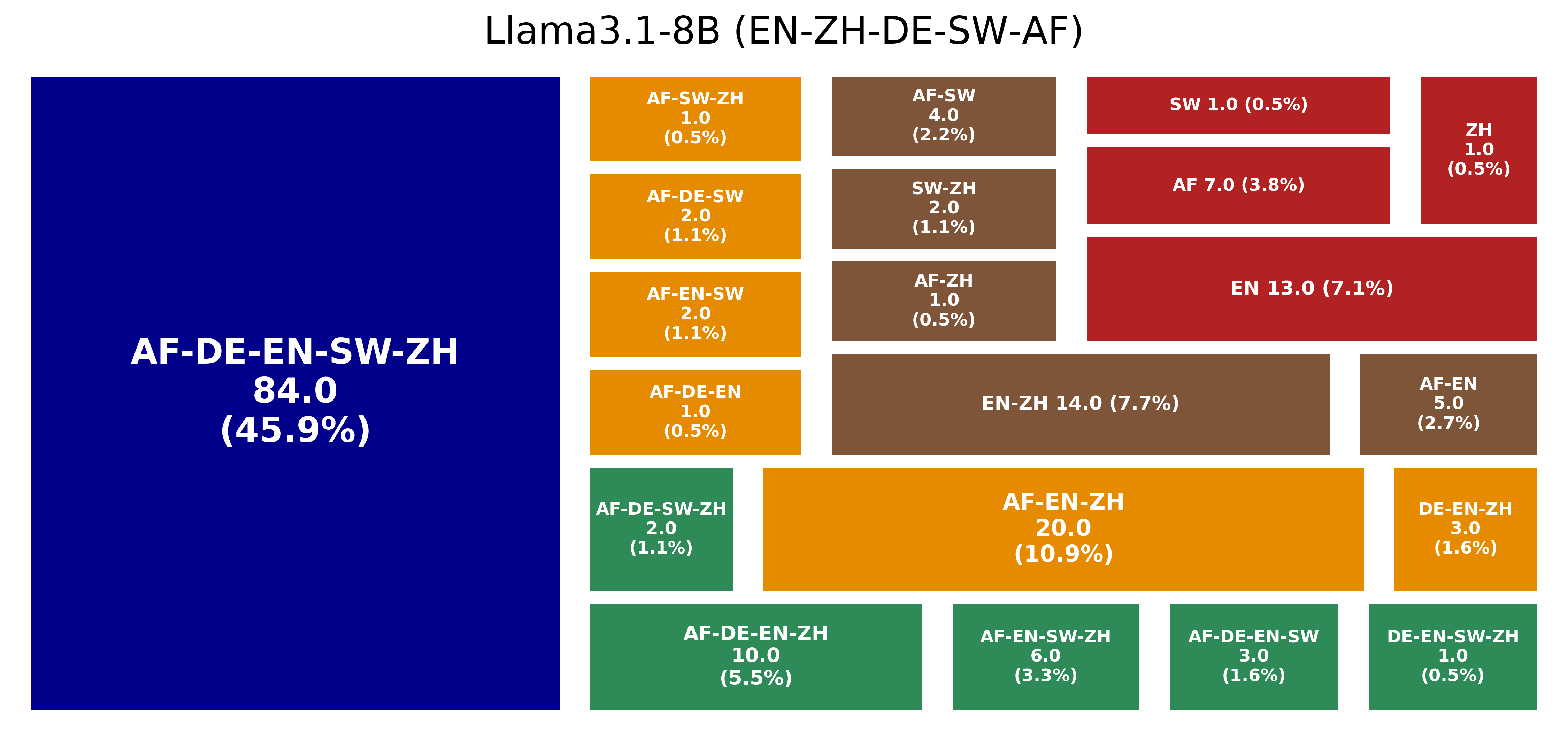}
\end{minipage}
\hfill
\begin{minipage}{0.48\linewidth} 
    \centering
    \includegraphics[width=0.875\linewidth]{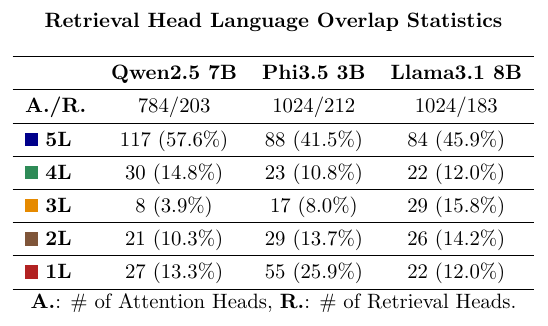}
\end{minipage}
\caption{
Language-specific (red) and shared (orange, green, blue) retrieval heads in three multilingual LMs for five languages - English (\texttt{en}), Chinese (\texttt{zh}), German (\texttt{de}), Swahili (\texttt{sw}), Afrikaans (\texttt{af}). $N$L represents intersection of retrieval head in $N$ languages, The size of each block is proportional to the number of retrieval heads that are \textcolor{color1L}{1-Language}, \textcolor{color2L}{2-Languages}, \textcolor{color3L}{3-Languages}, \textcolor{color4L}{4-Languages}, or \textcolor{color5L}{5-Languages}. Qwen2.5-7B exhibits strong sharing across languages (57.6\% 5L), Llama3.1-8B has high bilingual and trilingual head overlap (30\% 2L+3L), and Phi3.5-3B shows a measurably higher language-specific (mainly \texttt{en}-specific) behavior (25.9\% 1L).}
\label{fig:teaser}
\end{figure*}


To extend the NIAH task for different languages, we generate the data as follows (refer Figure~\ref{fig:niah_data_pipeline}). We first create a long-context text $c$ (haystack) using Wikipedia dumps of each language.\footnote{\url{https://en.wikipedia.org/wiki/Wikipedia:Database_download}} To create the needle for a particular language, we use Google Translate\footnote{\url{https://translate.google.com/}} to translate the same English needles used by \citet{wu2024retrieval} into the desired non-English language. Hence, while haystack changes for different languages based on Wikipedia dumps, the needle remains the same, just translated to the desired language. We then segment the haystack to match the desired context length and insert the needle at a controlled depth. This pipeline is fully automatic and readily scalable to other languages. Following \citet{wu2024retrieval}, we evaluate 600 examples, spanning \rebuttal{3 needles,} 20 context-length settings and 10 needle-depth percentages, and report each head’s mean score averaged over all configurations.

\paragraph{Models}
We use three popular open-source multilingual LLMs: Qwen2.5 7B Instruct \cite{qwen2025qwen25technicalreport}, Phi3.5 3B MiniInstruct \cite{abdin2024phi3technicalreporthighly}, and Llama3.1 8B Instruct \cite{grattafiori2024Llama}.\footnote{Given all are Instruct models, we omit the use of words `Instruct' \& `MiniInstruct' for readability.} 


\subsection{Results}

Figure \ref{fig:teaser} visualizes retrieval heads across five languages found in three models. We discuss their patterns below. 

\paragraph{Nearly half of retrieval heads are shared across languages} We color-code the retrieval heads by the number of languages that use that retrieval head. Across all three models, retrieval heads are frequently shared across languages. Roughly half of identified retrieval heads are shared among all five languages (57.6\% for Qwen2.5-7B, 45.9\% for Llama3.1-8B, and 41.5\% in Phi3.5-3B).


\paragraph{Differences among LLMs} We observe differences among language models. The proportion of bilingual and trilingual heads in Llama3.1-8B accounts for 30\% of all retrieval heads, more than twice the most abundant bilingual pairs in Qwen2.5-7B (14.2\%). Finally, Phi3.5-3B exhibits a disproportionately high number of \texttt{en}-specific retrieval heads (20.3\%), while other models have lower counts for \texttt{en}-specific heads (4.9\% and 7.1\% respectively). We hypothesize this English-inclined behavior of Phi3.5-3B to its small model size and a different synthetic post-training recipe, as further discussed in Appendix Table~\ref{appendix-tab:model_train_configs}. Our observations are consistent across increasing model scales and varying architectures (refer Appendix \ref{appendix:scale_size} and \ref{appendix:moe_vs_full} respectively).
 








\section{Retrieval-Transition heads}
\label{sec:RTH_Discovery}

\finaledits{Simply copying tokens from its context would be insufficient for handling complex tasks in multilingual models, as they operate within a target-language-agnostic concept space \cite{wendler2024Dollama,dumas-etal-2025-separating}. Prior studies have shown that this latent space is close to the English language due to high pretraining bias of English tokens \cite{wang2025lost,chen2025emergence,schut2025multilingual}. We modify the experimental setup of retrieval head detection to identify distinct heads, which we name as \textit{Retrieval-Transition heads (RTH)} that leverage this observation in multilingual LLMs}. In later section (\S~\ref{subsec:perf_RH_vs_RTH}), we evaluate how these heads compare to retrieval heads in impacting the performance of multilingual models.
\begin{figure*}[t!]
\centering
\includegraphics[width=1.0\textwidth]{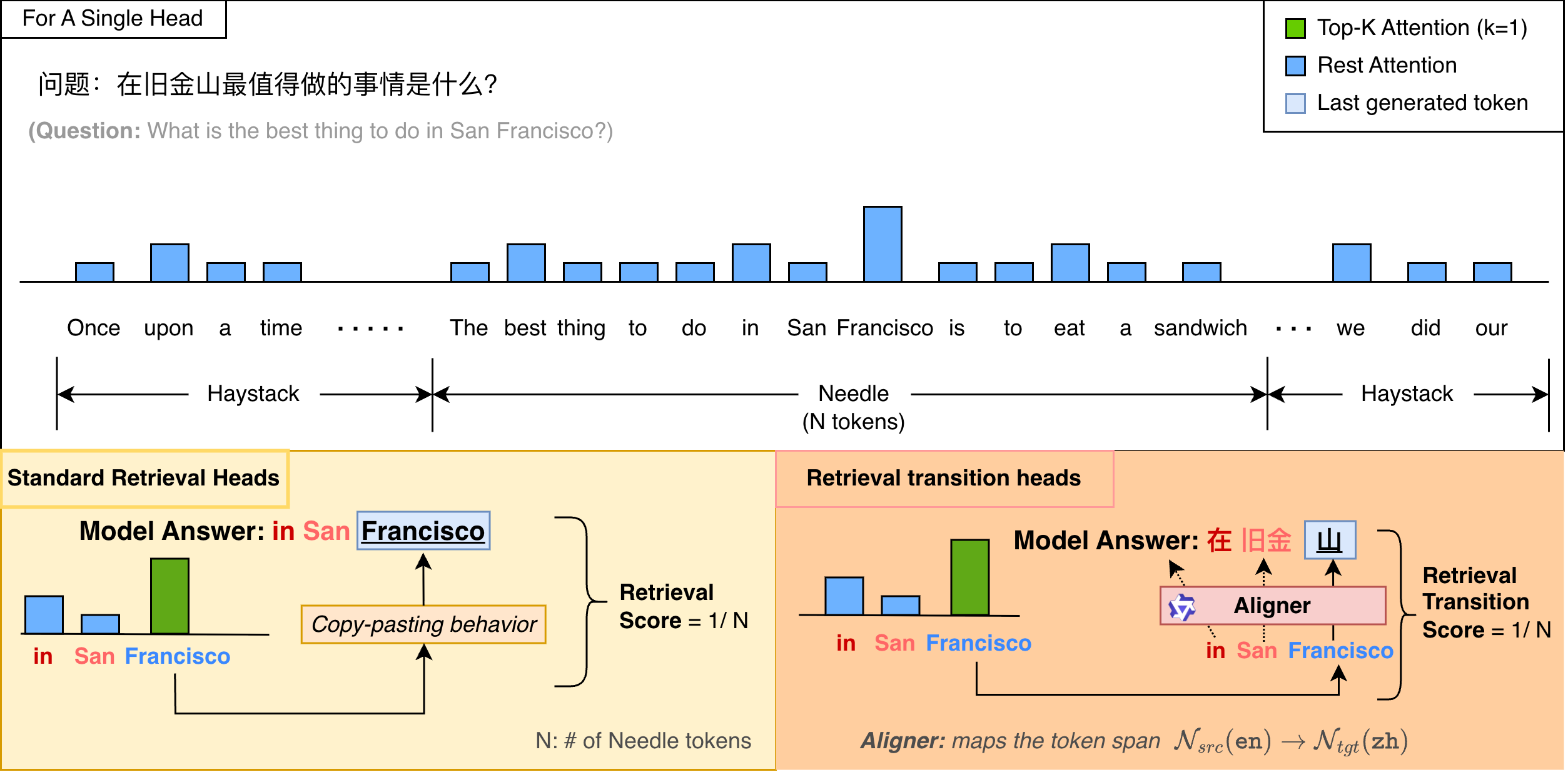}
\caption{Illustration of the retrieval-transition score (RTS). Needle tokens in the source language are first aligned to their target-language equivalents using a LM. The degree of overlap between the model’s generated tokens and these aligned target needles quantifies each attention head’s contribution to cross-lingual retrieval.}
\label{fig:pipe_trans}
\end{figure*}

\subsection{Experimental Setup} \label{sec-4:experimental-setup}




We introduce a variant of the NIAH task analogous to cross-lingual question answering. The haystack and needle are expressed in a language that approximates the concept space \(\mathcal{N}_{src}\) while the model is prompted to generate the output in a target language \(\mathcal{N}_{tgt}\) (see Figure~\ref{fig:pipe_trans}). 

Following prior work, we use English as the source language to approximate the model’s concept space, since the underlying latent token vocabulary is not directly observable \cite{wendler2024Dollama,dumas-etal-2025-separating}.
For models trained on more balanced multilingual corpora, such as the Qwen-2.5 family (see Appendix \ref{appendix:model_data}), we use Chinese as the source language to approximate the concept space when deriving retrieval-transition heads for English.


\paragraph{Retrieval transition score} 
At each decoding timestep $t$, this requires establishing a semantic mapping between the generated output token in the target language \(\mathcal{N}_{tgt}\) and its corresponding "needle" token in the source context \(\mathcal{N}_{src}\). We utilize \texttt{Qwen3-30B-A3B-Thinking-2507}\footnote{\url{https://huggingface.co/Qwen/Qwen3-30B-A3B-Thinking-2507}} to perform this alignment, ensuring that output tokens are correctly mapped to their source-language counterparts. We also evaluate an alternative LLM-based aligner (GPT 4.1) and find that the identified retrieval-transition heads are largely stable to the choice of aligner (Appendix~\ref{appendix:secD_2}). \camera{We further validate the aligner against human annotation of 1,644 en$\to$zh target subwords, obtaining step-level precision of 0.67, recall of 0.89 and F1 score of 0.76 (Appendix~\ref{app:aligner-validation}).}

The Retrieval Transition Score for a given attention head $h$ $(RTS_{h})$ is defined as:
\begin{equation}
    RTS_{h} = \frac{1}{N} \sum_{t=1}^{M} \mathcal{A}^t,
\end{equation}
\begin{equation}
    \text{where, } \mathcal{A}^t = \mathbbm{1} \left( \operatorname{\arg\max_{i}}a_{h,i}^t \in f^{-1}(\mathcal{O}) \right)
\end{equation}

\noindent Here, $RTS_h$ is the retrieval transition score for attention head $h$, $N$ is the number of tokens being evaluated in the transition sequence, $M$ is the max decoding limit, set to $M=50$, $\mathbbm{1}(\cdot)$ is the indicator function, $a_{h,i}^t$ denotes the attention weight assigned to input token $i$ by head $h$ at decoding timestep $t$, and $f^{-1}(\mathcal{O})$ represents the set of source-language needle token indices that map to the target-language output token $\mathcal{O}$ being decoded at time $t$ via the alignment function $f$. \camera{Same as $RS_h$ definition, we consider all heads with $RTH_h \ge 0.01$ as a Retrieval-Transition head.}
\begin{figure*}[t!]
\includegraphics[width=1.0\textwidth]{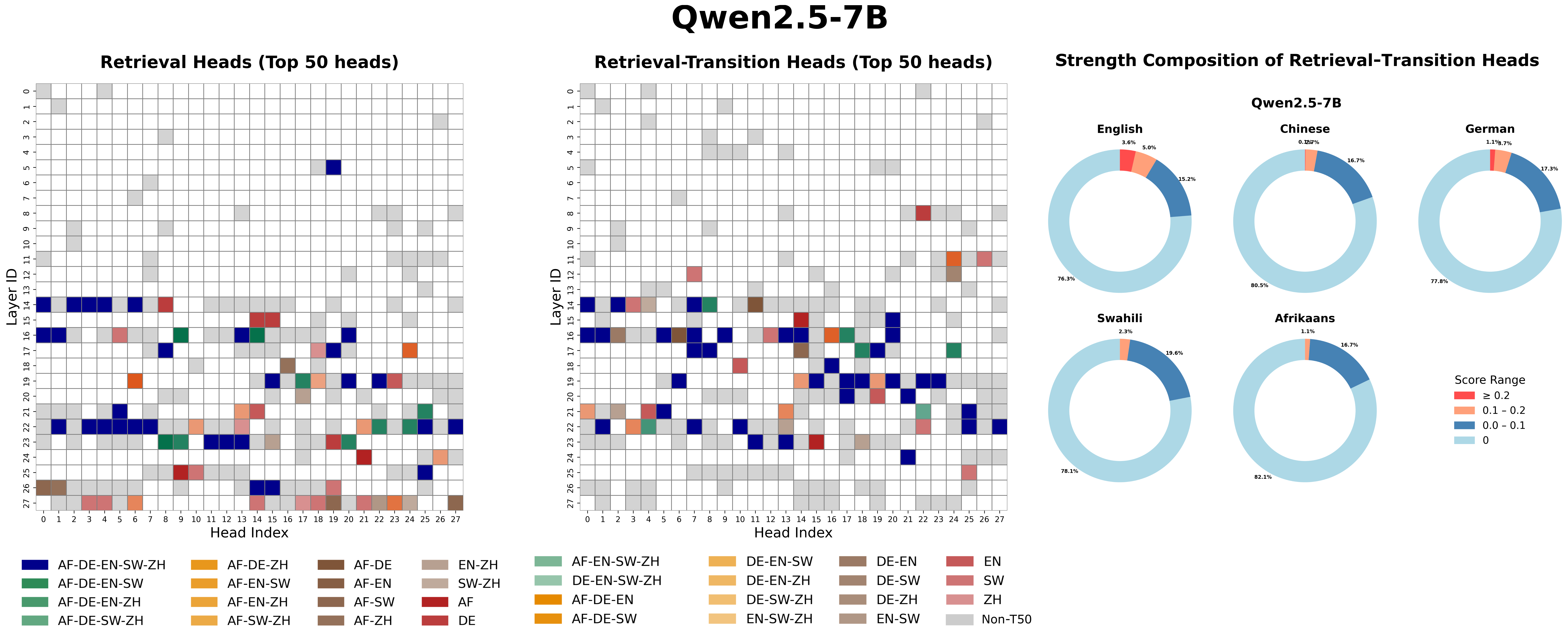}
\caption{Layer–head distributions of Retrieval Heads (RH) and Retrieval-Transition Heads (RTH) across languages for Qwen2.5-7B. Colored cells are part of Top-50 most prominent heads ranked by score (grey cells are RH/RTH but outside top-50). Among these colored heads, we find language-specific RH are dominant in the final layers while RTH are prominent in the middle layers. Like RH heads, RTH heads too show \textit{sparsity} with only 3-8\% of heads with an RTS score above 0.1}
\label{fig:loc_rh_rth_qwen}
\end{figure*}

\begin{figure}[h!]
    \centering
    \includegraphics[width=1.0\linewidth]{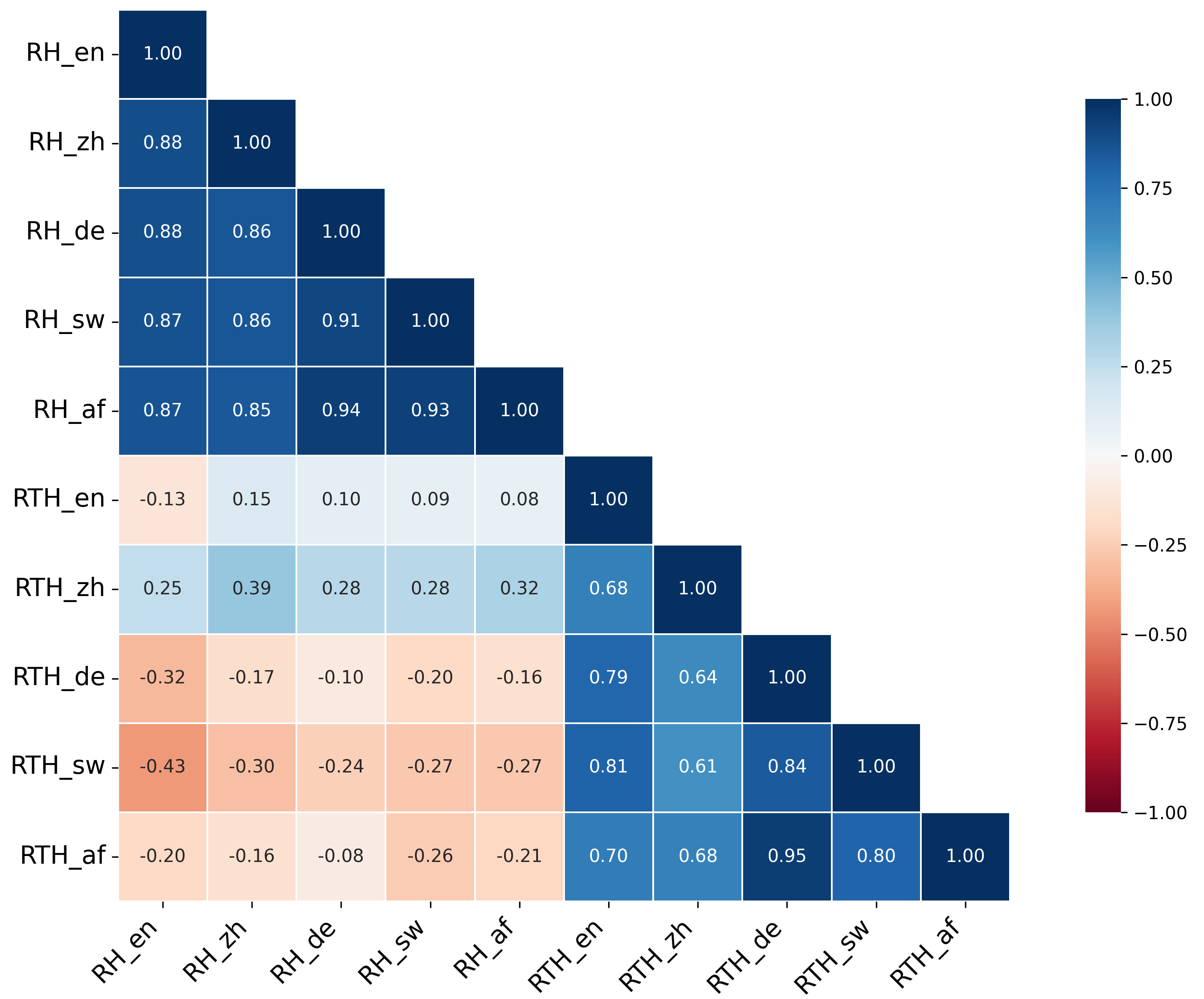}
    \caption{Spearman correlation score between top-50 most influential $RH(\ell)$ and $RTH(\ell)$ in Qwen2.5-7B. $RTH(\ell)$ heads share little correlation with $RH(\ell)$ but high correlations among other $RTH(\ell_0)$  and vice-versa}
    \label{fig:spearman_rh_rth}
\end{figure}

\paragraph{Models} Following previous section (c.f. \S~\ref{sec:3_models_and_langs}) we perform our experiments on Qwen2.5-7B, Phi3.5-3B, and Llama3.1-8B with English (\texttt{en}), German (\texttt{de}), Chinese (\texttt{zh}), and Swahili (\texttt{sw}) as target languages ($\mathcal{N}_{tgt}$). 
We use English as the source language ($\mathcal{N}_{src}$) for all experiments except for $RTH_{en}$ where we use Chinese.

\begin{table}[h!]
\centering
\small
\begin{tabular}{lcccc}
\toprule
Language & Top-15 & Top-25 & Top-50 & Top-100 \\
\midrule
EN & 2  & 10 & 26 & 63 \\
DE & 2  & 8  & 28 & 71 \\
ZH & 6  & 10 & 30 & 62 \\
SW & 2  & 10 & 27 & 57 \\
AF & 0  & 8  & 25 & 57 \\
\bottomrule
\end{tabular}
\caption{Number of intersecting top-$k$ heads $RH(\ell)$ and $RTH(\ell)$ for each language $\ell$ in Qwen2.5-7B. We find minimal overlap especially among the most influential $RH(\ell)$ and \rebuttal{$RTH(\ell)$ }suggesting their different functions.}
\label{tab:rh_rth_topk_intersection}
\end{table}

\begin{figure*}
\centering
\includegraphics[width=0.95\linewidth]{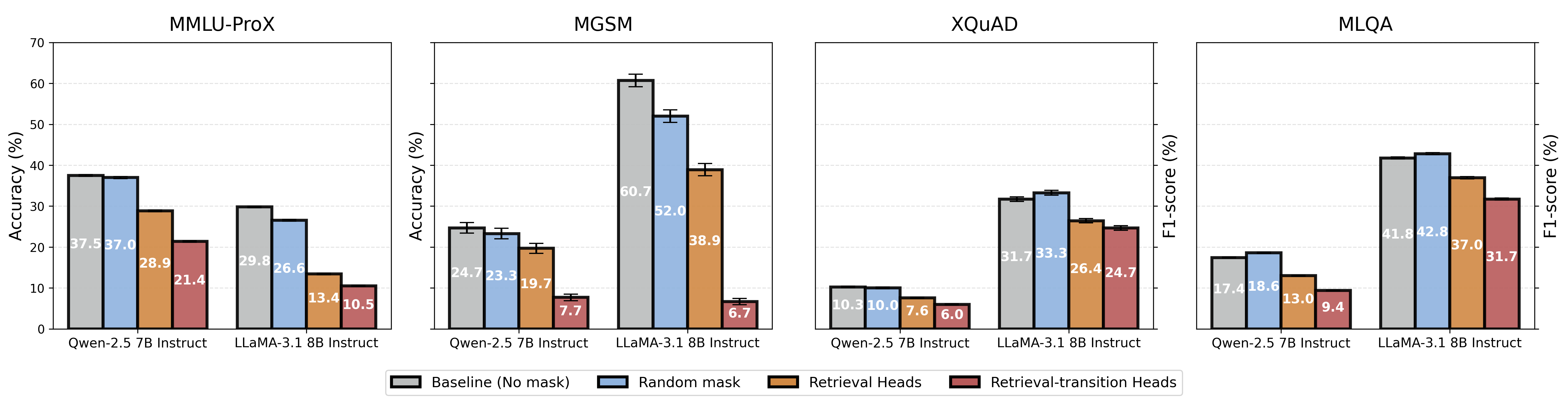}
\caption{\textbf{Impact of attention head masking ($k=25$) on downstream multilingual reasoning performance.} We compare the effects of masking random heads, Retrieval Heads (RH), and Retrieval-Transition Heads (RTH) on two models (Qwen-2.5-7B and Llama3.1-8B) averaged across all languages. While random masking has a negligible effect, masking RTHs (red) leads to a significantly more severe collapse in accuracy compared to RHs (orange), particularly on reasoning-heavy tasks like MGSM.}
\label{fig:masking_figure_4_bench}
\end{figure*}

\section{Analyzing Retrieval Transition Heads}
\label{sec:RTH_experiments}
\subsection{Comparing Retrieval Transition Heads with Retrieval Heads} \label{subsec:compare_RH_RTH} 
As seen in Table \ref{tab:rh_rth_topk_intersection}, the intersection between the RH and RTH is low for most prominent heads in both sets; for instance, only 2 out of the top 15 heads are common to both categories in English, German, and Swahili. Even among the top 50 heads, the overlap remains limited to approximately half of the heads.
Among the intersecting heads, we examine the Spearman rank correlation for top-50 most prominent heads in Qwen2.5-7B for RH and RTH across five languages: English (en), Chinese (zh), German (de), Swahili (sw), and Afrikaans (af). As illustrated in the global correlation matrix (Figure \ref{fig:spearman_rh_rth}), consistent clusters emerges within RH heads, and within RTH heads where coefficients remain high, typically ranging from 0.61 to 0.95. However, when comparing RTH with RH, even for the same language, we observe little to negative correlation. Figure \ref{fig:loc_rh_rth_qwen} (left) reports the location of RTHs in comparison to that of RHs for Qwen2.5-7B. The most prominent RTHs appear predominantly in the middle layers (14-23) while RHs appear mainly in the final layers (18 of top-50 heads in last 3 layers). \rebuttal{As shown in Appendix \ref{appendix:secD_1}, these characteristics are consistent across various model families, scales, and training recipes}. These results suggest that while within-language retrieval is unified, the transition from latent concepts to target languages relies on distinct, specialized circuits.

\begin{table*}[h!]
    \centering
    \definecolor{negcolor}{rgb}{0.70,0.13,0.13}
    \definecolor{poscolor}{rgb}{0.0,0.6,0.3}
    \definecolor{lightblue}{rgb}{0.9, 0.95, 1.0}

    \renewcommand{\arraystretch}{1.3}
    \footnotesize
    \setlength{\tabcolsep}{4pt}

    \resizebox{0.98\textwidth}{!}{
    \begin{tabular}{c|l|cccccc|cccc}
    \toprule
    \qquad & \multirow{2}{*}{\normalsize{\textbf{Heads}}} & \multicolumn{6}{c|}{\normalsize{\textbf{MMLU-ProX (Acc.)}}} & \multicolumn{4}{c}{\normalsize{\textbf{XQuAD (F1-score)}}} \\
    & & {\texttt{en}} & {\texttt{zh}} & {\texttt{de}} & {\texttt{sw}} & {\texttt{af}} & \textbf{Avg} & {\texttt{en}} & {\texttt{zh}} & {\texttt{de}} & \textbf{Avg} \\
    \midrule

    \rowcolor{lightgray}\cellcolor{white}\multirow{4}{*}{\rotatebox{90}{\textbf{Qwen2.5 7B}}}
    & None
    & 42.56 & 44.22 & 40.57 & 22.59 & 32.98 & 36.58
    & 15.72 & 0.82 & 14.28 & 10.27 \\

    & Random
    & 43.35$^{\color{poscolor}+0.79}$ & 42.66$^{\color{negcolor}-1.56}$ & 39.57$^{\color{negcolor}-1.00}$ & 22.32$^{\color{negcolor}-0.27}$ & 32.59$^{\color{negcolor}-0.39}$ & 36.10$^{\color{negcolor}-0.49}$
    & 15.94$^{\color{poscolor}+0.22}$ & 0.93$^{\color{poscolor}+0.11}$ & 13.23$^{\color{negcolor}-1.05}$ & 10.03$^{\color{negcolor}-0.24}$ \\

    & RH
    & 33.18$^{\color{negcolor}-9.38}$ & 33.42$^{\color{negcolor}-10.80}$ & 34.71$^{\color{negcolor}-5.86}$ & 14.18$^{\color{negcolor}-8.41}$ & 28.45$^{\color{negcolor}-4.53}$ & 28.79$^{\color{negcolor}-7.80}$
    & 13.44$^{\color{negcolor}-2.28}$ & 0.38$^{\color{negcolor}-0.44}$ & 8.97$^{\color{negcolor}-5.31}$ & 7.60$^{\color{negcolor}-2.68}$ \\

    & \cellcolor{lightblue}RTH
    & \cellcolor{lightblue}30.61$^{\color{negcolor}-11.95}$ & \cellcolor{lightblue}32.38$^{\color{negcolor}-11.84}$ & \cellcolor{lightblue}7.07$^{\color{negcolor}-33.50}$ & \cellcolor{lightblue}15.43$^{\color{negcolor}-7.16}$ & \cellcolor{lightblue}3.92$^{\color{negcolor}-29.06}$ & \cellcolor{lightblue}\textbf{17.88}$^{\color{negcolor}\textbf{-18.70}}$
    & \cellcolor{lightblue}11.80$^{\color{negcolor}-3.92}$ & \cellcolor{lightblue}0.36$^{\color{negcolor}-0.46}$ & \cellcolor{lightblue}5.78$^{\color{negcolor}-8.50}$ & \cellcolor{lightblue}\textbf{5.98}$^{\color{negcolor}\textbf{-4.29}}$ \\

    \midrule

    \rowcolor{lightgray}\cellcolor{white}\multirow{4}{*}{\rotatebox{90}{\textbf{Llama3.1 8B}}}
    & None
    & 43.59 & 31.41 & 30.89 & 13.27 & 27.38 & 29.31
    & 35.26 & 28.70 & 31.10 & 31.69 \\

    & Random
    & 41.70$^{\color{negcolor}-1.89}$ & 28.94$^{\color{negcolor}-2.47}$ & 22.87$^{\color{negcolor}-8.02}$ & 12.71$^{\color{negcolor}-0.56}$ & 21.39$^{\color{negcolor}-5.99}$ & 25.52$^{\color{negcolor}-3.79}$
    & 37.18$^{\color{poscolor}+1.92}$ & 29.57$^{\color{poscolor}+0.87}$ & 33.03$^{\color{poscolor}+1.93}$ & 33.26$^{\color{poscolor}+1.57}$ \\

    & RH
    & 17.03$^{\color{negcolor}-26.56}$ & 20.70$^{\color{negcolor}-10.71}$ & 12.65$^{\color{negcolor}-18.24}$ & 3.27$^{\color{negcolor}-10.00}$ & 6.81$^{\color{negcolor}-20.57}$ & 12.09$^{\color{negcolor}-17.22}$
    & 30.35$^{\color{negcolor}-4.91}$ & 22.84$^{\color{negcolor}-5.86}$ & 26.12$^{\color{negcolor}-4.98}$ & 26.44$^{\color{negcolor}-5.25}$ \\

    & \cellcolor{lightblue}RTH
    & \cellcolor{lightblue}11.42$^{\color{negcolor}-32.17}$ & \cellcolor{lightblue}24.76$^{\color{negcolor}-6.65}$ & \cellcolor{lightblue}2.35$^{\color{negcolor}-28.54}$ & \cellcolor{lightblue}3.50$^{\color{negcolor}-9.77}$ & \cellcolor{lightblue}1.03$^{\color{negcolor}-26.35}$ & \cellcolor{lightblue}\textbf{8.61}$^{\color{negcolor}\textbf{-20.70}}$
    & \cellcolor{lightblue}19.66$^{\color{negcolor}-15.60}$ & \cellcolor{lightblue}24.99$^{\color{negcolor}-3.71}$ & \cellcolor{lightblue}29.38$^{\color{negcolor}-1.72}$ & \cellcolor{lightblue}\textbf{24.68}$^{\color{negcolor}\textbf{-7.01}}$ \\

    \bottomrule
    \end{tabular}
    }

    \vspace{8pt}

    \resizebox{0.98\textwidth}{!}{
    \begin{tabular}{c|l|ccccc|cccc}
    \toprule
    \qquad & \multirow{2}{*}{\normalsize{\textbf{Heads}}} & \multicolumn{5}{c|}{\normalsize{\textbf{MGSM (Acc.)}}} & \multicolumn{4}{c}{\normalsize{\textbf{MLQA (F1-score)}}} \\
    & & {\texttt{en}} & {\texttt{zh}} & {\texttt{de}} & {\texttt{sw}} & \textbf{Avg} & {\texttt{en}} & {\texttt{zh}} & {\texttt{de}} & \textbf{Avg} \\
    \midrule

    \rowcolor{lightgray}\cellcolor{white}\multirow{4}{*}{\rotatebox{90}{\textbf{Qwen2.5 7B}}}
    & None
    & 33.6 & 31.6 & 27.2 & 6.4 & 24.70
    & 17.26 & 18.93 & 16.16 & 17.45 \\

    & Random
    & 32.0$^{\color{negcolor}-1.6}$ & 27.2$^{\color{negcolor}-4.4}$ & 30.0$^{\color{poscolor}+2.8}$ & 4.0$^{\color{negcolor}-2.4}$ & 23.30$^{\color{negcolor}-1.40}$
    & 18.61$^{\color{poscolor}+1.35}$ & 19.88$^{\color{poscolor}+0.95}$ & 17.29$^{\color{poscolor}+1.13}$ & 18.59$^{\color{poscolor}+1.14}$ \\

    & RH
    & 31.2$^{\color{negcolor}-2.4}$ & 22.8$^{\color{negcolor}-8.8}$ & 21.2$^{\color{negcolor}-6.0}$ & 3.6$^{\color{negcolor}-2.8}$ & 19.70$^{\color{negcolor}-5.00}$
    & 15.05$^{\color{negcolor}-2.21}$ & 12.74$^{\color{negcolor}-6.19}$ & 11.20$^{\color{negcolor}-4.96}$ & 13.00$^{\color{negcolor}-4.45}$ \\

    & \cellcolor{lightblue}RTH
    & \cellcolor{lightblue}24.0$^{\color{negcolor}-9.6}$ & \cellcolor{lightblue}2.8$^{\color{negcolor}-28.8}$ & \cellcolor{lightblue}3.2$^{\color{negcolor}-24.0}$ & \cellcolor{lightblue}0.8$^{\color{negcolor}-5.6}$ & \cellcolor{lightblue}\textbf{7.70}$^{\color{negcolor}\textbf{-17.00}}$
    & \cellcolor{lightblue}13.27$^{\color{negcolor}-3.99}$ & \cellcolor{lightblue}8.29$^{\color{negcolor}-10.64}$ & \cellcolor{lightblue}6.54$^{\color{negcolor}-9.62}$ & \cellcolor{lightblue}\textbf{9.37}$^{\color{negcolor}\textbf{-8.08}}$ \\

    \midrule

    \rowcolor{lightgray}\cellcolor{white}\multirow{4}{*}{\rotatebox{90}{\textbf{Llama3.1 8B}}}
    & None
    & 73.2 & 62.4 & 59.2 & 48.0 & 60.70
    & 39.05 & 49.49 & 36.76 & 41.77 \\

    & Random
    & 69.2$^{\color{negcolor}-4.0}$ & 55.6$^{\color{negcolor}-6.8}$ & 56.0$^{\color{negcolor}-3.2}$ & 27.2$^{\color{negcolor}-20.8}$ & 52.00$^{\color{negcolor}-8.70}$
    & 40.29$^{\color{poscolor}+1.24}$ & 49.49$^{\phantom{+}0.00}$ & 38.61$^{\color{poscolor}+1.85}$ & 42.80$^{\color{poscolor}+1.03}$ \\

    & RH
    & 56.0$^{\color{negcolor}-17.2}$ & 42.0$^{\color{negcolor}-20.4}$ & 38.8$^{\color{negcolor}-20.4}$ & 18.8$^{\color{negcolor}-29.2}$ & 38.90$^{\color{negcolor}-21.80}$
    & 34.10$^{\color{negcolor}-4.95}$ & 44.25$^{\color{negcolor}-5.24}$ & 32.49$^{\color{negcolor}-4.27}$ & 36.95$^{\color{negcolor}-4.82}$ \\

    & \cellcolor{lightblue}RTH
    & \cellcolor{lightblue}3.6$^{\color{negcolor}-69.6}$ & \cellcolor{lightblue}5.6$^{\color{negcolor}-56.8}$ & \cellcolor{lightblue}15.2$^{\color{negcolor}-44.0}$ & \cellcolor{lightblue}2.4$^{\color{negcolor}-45.6}$ & \cellcolor{lightblue}\textbf{6.70}$^{\color{negcolor}\textbf{-54.00}}$
    & \cellcolor{lightblue}21.90$^{\color{negcolor}-17.15}$ & \cellcolor{lightblue}38.92$^{\color{negcolor}-10.57}$ & \cellcolor{lightblue}34.41$^{\color{negcolor}-2.35}$ & \cellcolor{lightblue}\textbf{31.74}$^{\color{negcolor}\textbf{-10.02}}$ \\

    \bottomrule
    \end{tabular}
    }

    \caption{\textbf{Per-language performance degradation under top-$k$ ($k=25$) head masking.} Accuracy (MMLU-ProX, MGSM) and F1-scores (MLQA, XQuAD) are reported for Qwen-2.5-7B and Llama3.1 8B. The \textbf{Avg} column reports the mean across the evaluated languages, with the colored superscript denoting the absolute change relative to the no-masking baseline. Results highlight that RTH masking consistently induces the largest performance drops across high-resource (en, zh, de) and often in low-resource (sw) languages. Crucially, English-language tasks are also affected when masking RTH, suggesting that the latent reasoning mechanism in multilingual language models is invariant to the target language.}
    \label{tab:masking_results_with_all_lang_k25}
\end{table*}

\subsection{Its impact on task performance} \label{subsec:perf_RH_vs_RTH}

\paragraph{Evaluation Setting} \rebuttal{Following \citet{wu2024retrieval}, we quantify the causal impact of specific heads by perturbing their pre-softmax attention logits during the forward pass; a formal description of this intervention is in Appendix \ref{appendix:causal_method}.} We use a monolingual setting for each language $(L)$, where both the source context and the expected output are in $L$. \camera{In this way we can show how masking RTH can affect performance when the task requires no explicit cross-lingual mapping.} We evaluate three masking strategies: (i) masking \textit{k} randomly selected heads, (ii) masking the top-\textit{k} $RH_{L}$, and (iii) masking the top-\textit{k} $RTH_{L}$. In all experiments, we set $\textit{k}=25$ and use chain-of-thought prompt template \cite{eval-harness}. Among the evaluated benchmarks, Afrikaans appears only in MMLU-ProX.

\paragraph{Evaluation Benchmark} {We study four multilingual benchmarks described below. 
MMLU-ProX \cite{xuan-etal-2025-mmlu} is a multilingual, reasoning-centric multiple-choice QA benchmark that requires complex, multi-step reasoning and covers 29 languages. MGSM \cite{shi2022language} is an expertly annotated multilingual extension of the GSM-8k \cite{cobbe2021training} dataset and consists of arithmetic word problems to evaluate numerical and logical reasoning across multiple languages while controlling for problem structure. MLQA \cite{lewis2020mlqa} is a cross-lingual extractive question answering benchmark that tests multilingual reading comprehension by requiring models to locate answer spans in passages written in different languages. XQuAD \cite{artetxe2020cross} is a multilingual dataset measuring zero-shot transfer, where models trained on English must answer extractive questions in target languages.}

\begin{table*}[t]
\centering
\small
\setlength{\tabcolsep}{5pt}
\begin{tabular}{c|c|c|ccc|ccc}
\hline
\textbf{Dataset} & \textbf{Setting} & \multicolumn{1}{c|}{\textbf{\# Samples}} &
\multicolumn{3}{c|}{\textbf{RH Failure Modes}} &
\multicolumn{3}{c}{\textbf{RTH Failure Modes}} \\
 &  &  & LoC & RF & Other & LoC & RF & Other \\
\hline
\multirow{4}{*}{\textbf{MMLU-ProX}}
& en & 1454 & 150 \pct{10.32} & 1269 \pct{87.28} & 34 \pct{2.34} 
& 446 \pct{30.70} & 997 \pct{68.62} & 10 \pct{0.69} \\
& zh & 1311 & 89 \pct{6.79}  & 1202 \pct{91.69} & 18 \pct{1.37} 
& 303 \pct{23.15} & 1003 \pct{76.62} & 3 \pct{0.23} \\
& de & 1738 & 230 \pct{13.23} & 1494 \pct{85.96} & 10 \pct{0.58} 
& 1116 \pct{64.29} & 618 \pct{35.60} & 2 \pct{0.12} \\
& sw & 890  & 749 \pct{84.16} & 141 \pct{15.84} & 0 \pct{0.00}  
& 824 \pct{92.58} & 66 \pct{7.42} & 0 \pct{0.00} \\
& af & 1707 & 454 \pct{26.60} & 1236 \pct{72.41} & 17 \pct{1.00}
  & 1415 \pct{82.89} & 291 \pct{17.05} & 1 \pct{0.06} \\
\hline
\multirow{4}{*}{\textbf{MGSM}}
& en & 52 & 0 \pct{0.00}  & 52 \pct{100.00} & 0 \pct{0.00}  
& 6 \pct{11.54}  & 45 \pct{86.54} & 1 \pct{1.92} \\
& zh & 62 & 3 \pct{4.84}  & 54 \pct{87.10}  & 5 \pct{8.06}  
& 6 \pct{9.68}   & 56 \pct{90.32} & 0 \pct{0.00} \\
& de & 61 & 7 \pct{11.48} & 54 \pct{88.52}  & 0 \pct{0.00}  
& 18 \pct{29.51} & 42 \pct{68.85} & 1 \pct{1.64} \\
& sw & 22 & 10 \pct{45.45} & 12 \pct{54.55}  & 0 \pct{0.00}  
& 20 \pct{90.91} & 2 \pct{9.09}   & 0 \pct{0.00} \\
\hline
\end{tabular}
\caption{\textbf{Quantitative breakdown of failure modes}—Loss of Coherence (LoC), Failure to Retrieve (RF), and Other—across MMLU-ProX and MGSM datasets for English (en), Chinese (zh), German (de), Swahili (sw) and Afrikaan (af) on Qwen2.5-7B. Each cell reports \# (count) \pct{\% (percentage)}. The table compares failure modes under RH vs. RTH masking on samples that flip in both settings (No-mask$\rightarrow$RH and No-mask$\rightarrow$RTH). Notably, masking RTHs consistently yields a higher proportion of LoC failures compared to RH masking across most languages, with German exhibiting the highest spike in loss of coherence for MMLU-ProX.}
\label{tab:failure_mode_counts_with_pct_rh_rth}
\end{table*}

\paragraph{Results} \rebuttal{Figure~\ref{fig:masking_figure_4_bench} summarizes the effect of masking different attention head sets, averaged across languages, for the Qwen2.5-7B and Llama3.1-8B models across the MMLU-ProX, MGSM, MLQA, and XQuAD benchmarks. We observe that masking $k=25$ random attention heads has a negligible effect on average, aside from a few low-resource languages, on downstream performance. Masking an equivalent number of Retrieval Heads (RH) leads to a measurable decline and masking Retrieval-Transition Heads (RTH) leads to even further drops. A striking example is found in mathematical reasoning (MGSM), where performance drops by 54 points for Llama3.1-8B and 17 points for Qwen2.5-7B.} Overall, these aggregate results indicate that retrieval-transition heads are more critical than retrieval heads for maintaining multilingual reasoning capabilities.

\noindent {Table~\ref{tab:masking_results_with_all_lang_k25} provides a granular, per-language breakdown across the four benchmarks. Aside from Swahili (\texttt{sw}) in MMLU-ProX, the disparity in performance loss between RH and RTH masking is significant across all languages. Swahili's outlier behavior can be attributed to the model's weak baseline performance on the language; even when Retrieval Heads are masked, the model’s primary failure mode is a loss of linguistic coherence rather than a failure to retrieve information (refer \S~\ref{subsec:qualitative}). Notably, masking RTHs triggers a severe collapse even in English-language tasks, far exceeding the degradation seen when masking RHs or random heads. \camera{We further experiment by varying the hyperparameter $k \in \{15, 50\}$. While our core finding remains consistent across values---RTH masking causes a larger performance drop than RH or Random masking---we observe diminishing differences at higher $k$. At $k \ge 50$, retrieval failures (missing information relevant to the question) drive most of the degradation by disrupting the model's chain-of-thought, largely overwhelming the effect of a weakened transition from latent representations to language-specific decoding. Refer to Appendix \ref{appendix:secE_1} for detailed results.} 

\subsection{Qualitative Analysis} \label{subsec:qualitative}
We conduct a qualitative error analysis to better understand the failures induced by masking, identifying two dominant failure modes: loss of coherence (LoC) and failure to retrieve (RF). In the LoC mode, masking RTHs fundamentally disrupts generation, causing the model to frequently collapse into nonsensical outputs, repetition, or looping behaviors. In contrast, in the RF mode, the model maintains fluency but does not retrieve the key information required to answer correctly—even when that information resides within the provided context or the model’s own chain-of-thought.

To quantify these failure modes, we evaluate Qwen2.5-7B at $k=25$ using Qwen3-235B-A22B \cite{yang2025qwen3technicalreport} as an LLM-as-a-judge (see Figure~\ref{fig:prompt_for_classifcation} for the evaluation prompt). We analyze "flipped" samples, which we define as instances answered correctly under No-Mask but incorrectly under both RH-Mask and RTH-Mask. \camera{To validate this judge against human annotation, one author independently re-annotated a blind, stratified sample of 50 of these instances using the same rubric. The human annotator and the judge assigned the same failure mode on 78.0\% of items. Cohen's $\kappa$ is 0.633, which falls in the substantial agreement band \cite{landis1977}. Appendix~\ref{appendix:qualitative_human_validation} gives the annotation protocol, sample composition, and full confusion matrix.}

Table~\ref{tab:failure_mode_counts_with_pct_rh_rth} shows the breakdown of different failure modes for two masking condition. The proportion of LoC failures is notably higher when masking RTHs compared to masking RHs. Furthermore, we observe cross-lingual and cross-dataset variations. On MMLU-ProX, Swahili exhibits a unique error profile under RH-Mask compared to other languages, with generation failures occurring substantially more frequently than retrieval failures. Under RTH-Mask, however, lack of coherence failures mode increases substantially across all evaluated languages. We observe a similar trend on MGSM, where masking RTHs consistently induces higher rates of LoC as compared to masking RH. The primary cross-dataset discrepancy lies in how Swahili responds to RH-Mask: while RH masking triggers a high volume of generation failures on MMLU-ProX, this degradation is significantly less pronounced on MGSM. This suggests that when evaluating low-resource languages on tasks requiring generative synthesis with minimal fact retrieval (such as MGSM), RTHs are likely more critical than RHs, especially in comparison to tasks that demand precise information recovery from either the local context or parametric memory (such as MMLU-ProX). In Appendix~\ref{appendix:sec_Qualitative}, we present several illustrative MMLU-ProX examples that highlight these failure modes: the model answers correctly under No-Mask and RH-Mask, but breaks down under RTH-Mask.

\section{Conclusion and Future Work}
\rebuttal{
This paper identifies Retrieval-Transition Heads (RTHs), a specialized sparse set of attention heads within multilingual LLMs responsible for bridging the gap between latent reasoning and target-language generation. Unlike retrieval heads that implement a copy-paste mechanism, RTHs bridge the target-language independent latent space to the final output. We demonstrate that RTHs are functionally indispensable for complex tasks: masking them triggers a catastrophic performance collapse on multilingual reasoning benchmarks. These findings have implications for efficiency methods. Recent frameworks for KV-cache compression \cite{fu2025not,tang2024razorattention}, efficient inference \cite{xiao2024duoattention}, and context re-ranking \cite{zhang2025query} rely on heuristics that prioritize retrieval behavior. Our results suggest these methods risk pruning the heads required for multilingual reasoning. We propose that future "transition-aware" strategies must incorporate RTH identification to preserve the specific heads responsible for translating latent context into linguistic fluency, thereby preventing the reasoning degradation that standard pruning would otherwise cause in multilingual settings.}
\\ 
\section*{Limitations}
First, the Retrieval Transition Score (RTS) approximates the language-agnostic concept space using English as a proxy; while effective due to inherent model biases, this results in RTHs that are less sharply defined than standard retrieval heads. \camera{Also, while this study contrasts between RH and RTH, we also see significant intersection among them (Table \ref{tab:rh_rth_topk_intersection}), which warrants future investigation on their role in multilingual retrieval. Another limitation of this study is its exclusive focus on attention heads. While heads are accessible units for causal masking, they represent only one part of the model's reasoning circuitry. Future work should explore the interplay between these heads, MLP 'shared neurons', and residual stream activations to provide a more holistic understanding of the latent-to-target language transition.} Finally, as models are trained with more balanced multilingual data, the single-language approximation (e.g. English) becomes typologically distant from the model's latent space reducing the discriminative signal of the method. Future research should transcend this reliance on pivot languages by developing systems that learn to mimic true latent tokens. By generating a model-specific latent vocabulary, researchers can move toward identifying transition mechanisms that reflect the model's internal conceptual structures rather than cross-lingual projections.

\section*{Acknowledgment}
 This work was supported in part through the NYU IT High Performance Computing resources, services, and staff expertise. The work is partially funded by NSF CAREER award 2443271 and NSF award RI-2521091.
\bibliography{anthology,custom}

\onecolumn  
\clearpage

\appendix
\renewcommand{\thefigure}{A\arabic{figure}}
\renewcommand{\thetable}{A\arabic{table}}

\setcounter{figure}{0}
\setcounter{table}{0}


\section{Multilingual Needles and Questions}
\label{sec:appendix_needles}

\camera{To enable community review of needle translation quality and support reproducibility, we present the evaluation needles and corresponding questions across Afrikaans, German, English, Swahili, and Chinese below.}

\begin{figure}[h!]
    \centering
    \includegraphics[width=0.9\linewidth]{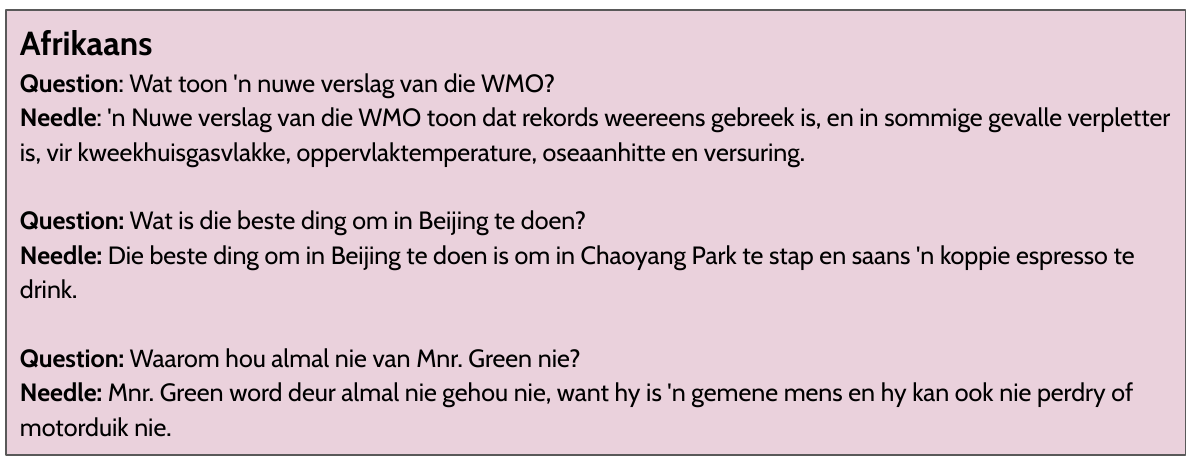}
    \caption{Afrikaans evaluation questions and needles.}
    \label{fig:afrikaans_needles}
\end{figure}

\begin{figure}[h!]
    \centering
    \includegraphics[width=0.9\linewidth]{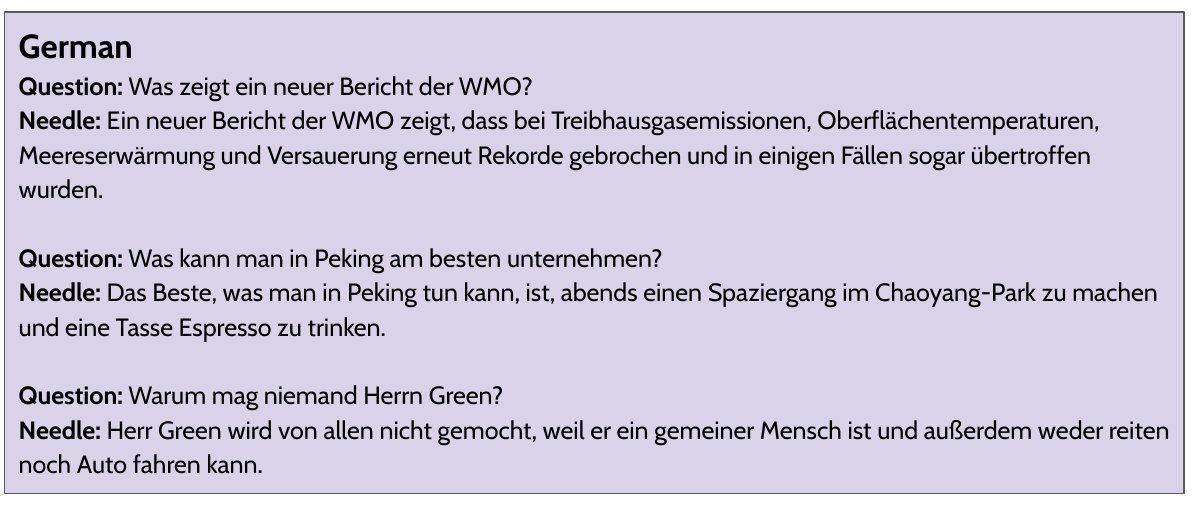}
    \caption{German evaluation questions and needles.}
    \label{fig:german_needles}
\end{figure}

\begin{figure}[h!]
    \centering
    \includegraphics[width=0.9\linewidth]{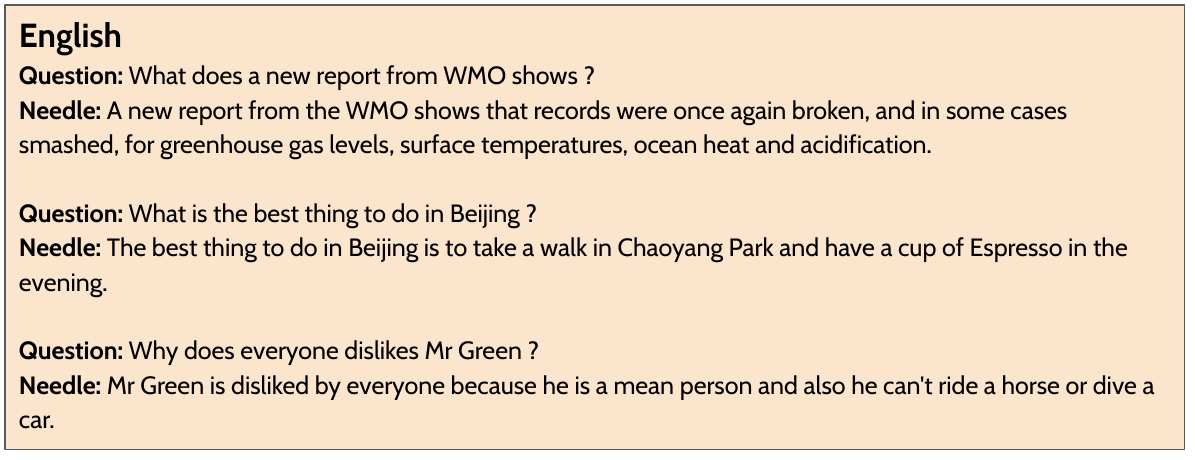}
    \caption{English evaluation questions and needles.}
    \label{fig:english_needles}
\end{figure}

\begin{figure}[h]
    \centering
    \includegraphics[width=0.9\linewidth]{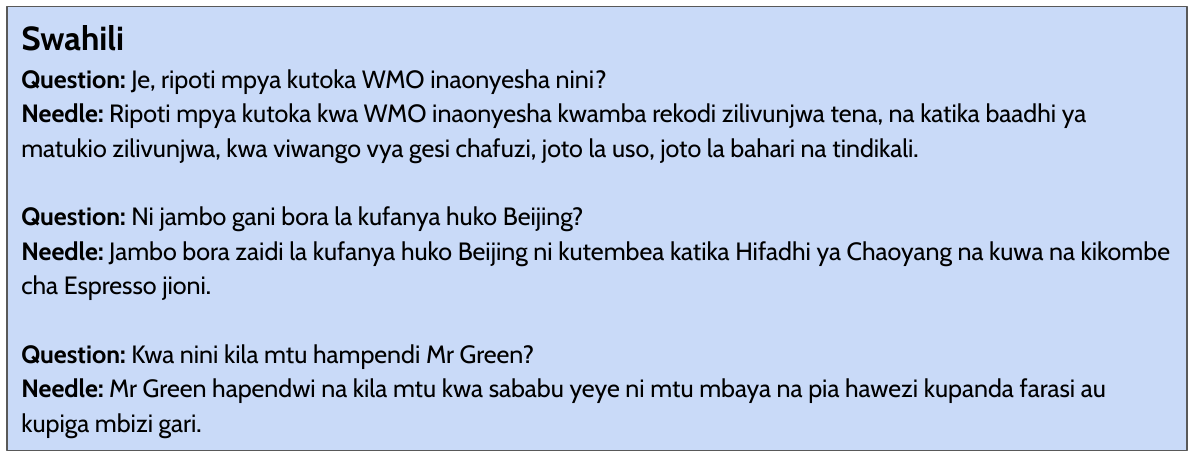}
    \caption{Swahili evaluation questions and needles.}
    \label{fig:swahili_needles}
\end{figure}

\begin{figure}[h!]
    \centering
    \includegraphics[width=0.9\linewidth]{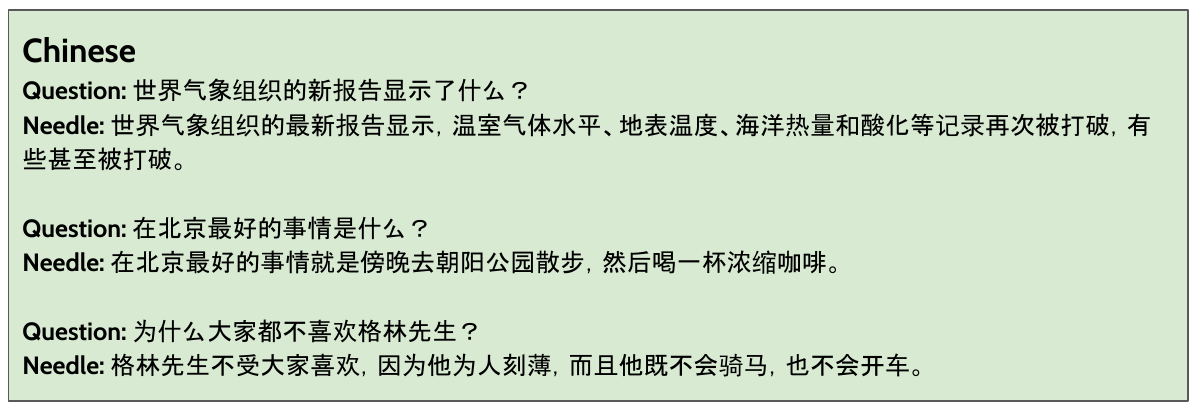}
    \caption{Chinese evaluation questions and needles.}
    \label{fig:chinese_needles}
\end{figure}

\section{Effect of model sizes, architecture, and low-resource languages to retrieval head analysis.}

\subsection{Effect of scaling model size on Qwen2.5}\label{appendix:scale_size}
\begin{figure}[h!]
\centering
\includegraphics[width=0.47\linewidth]{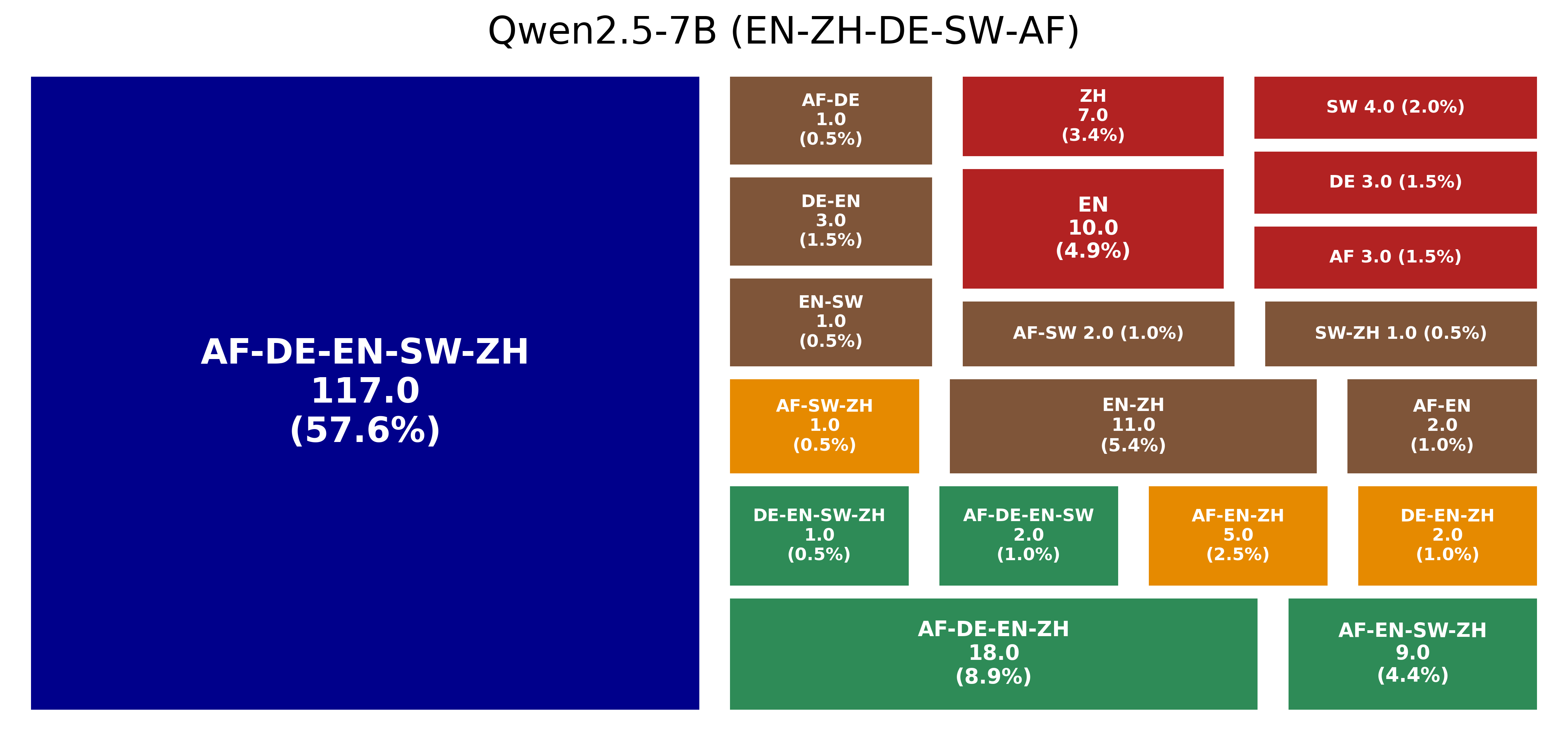}\hfill
\includegraphics[width=0.47\linewidth]{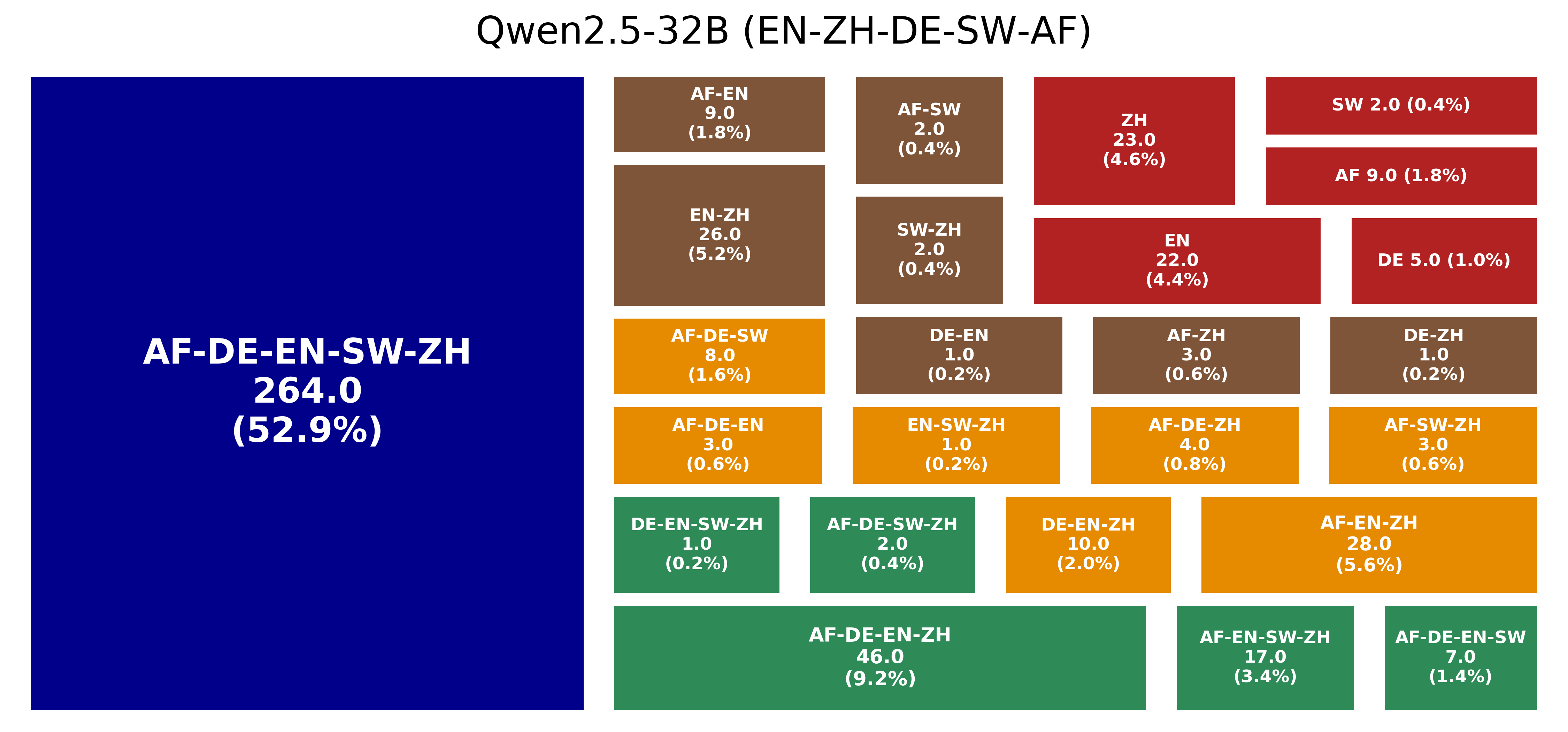}
\caption{To assess the effect of model scaling, we evaluate Qwen2.5-14B Instruct and Qwen2.5-32B Instruct. Despite the larger parameter count, the proportions of shared and language-exclusive attention heads remain largely consistent.}
\label{fig:qwen_scale}
\end{figure}

\begin{figure}[h!]
\centering
\includegraphics[width=0.47\linewidth]{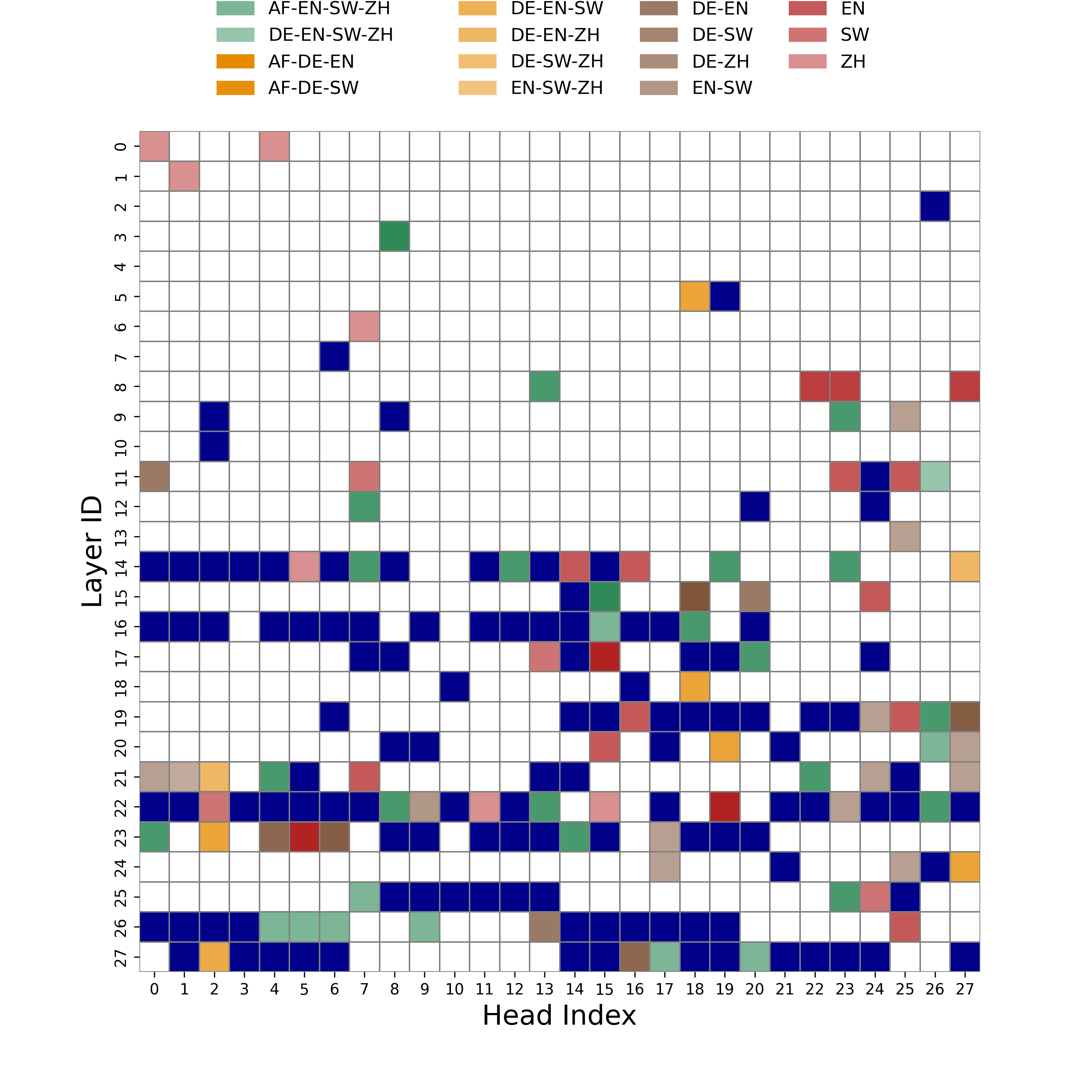}\hfill
\includegraphics[width=0.47\linewidth]{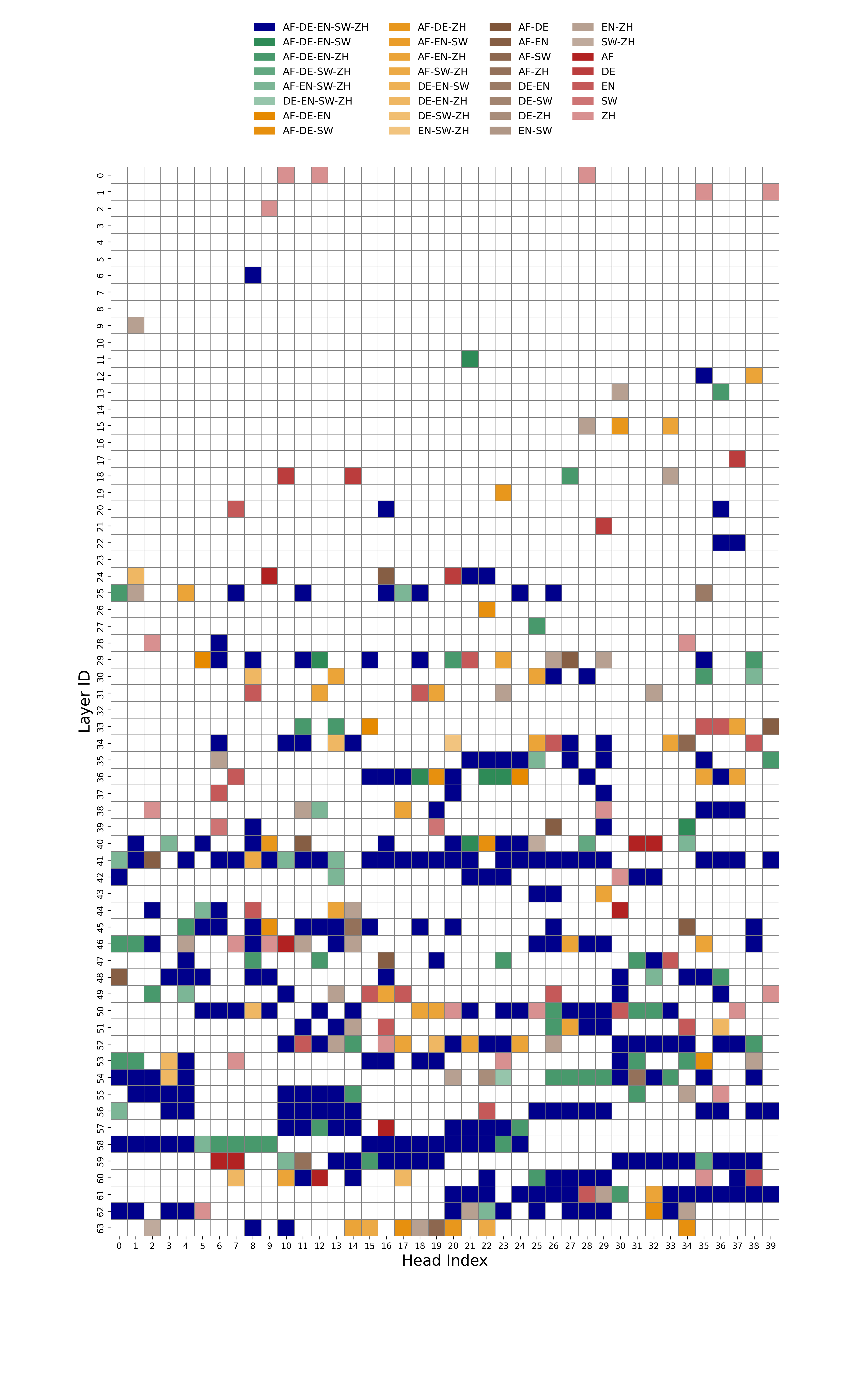}
\caption{To assess how scaling affects the placement of shared and language-exclusive attention heads, we evaluate Qwen2.5-14B Instruct and Qwen2.5-32B Instruct. The spatial distribution of these head types remains largely consistent across scales.}
\label{fig:qwen_scale_location}
\end{figure}

\vspace{3cm}

\subsection{Comparing Mixture-of-Experts to fully activated model architecture on Phi3.5} \label{appendix:moe_vs_full}

\begin{figure}[H]
\centering
\includegraphics[width=0.47\linewidth]{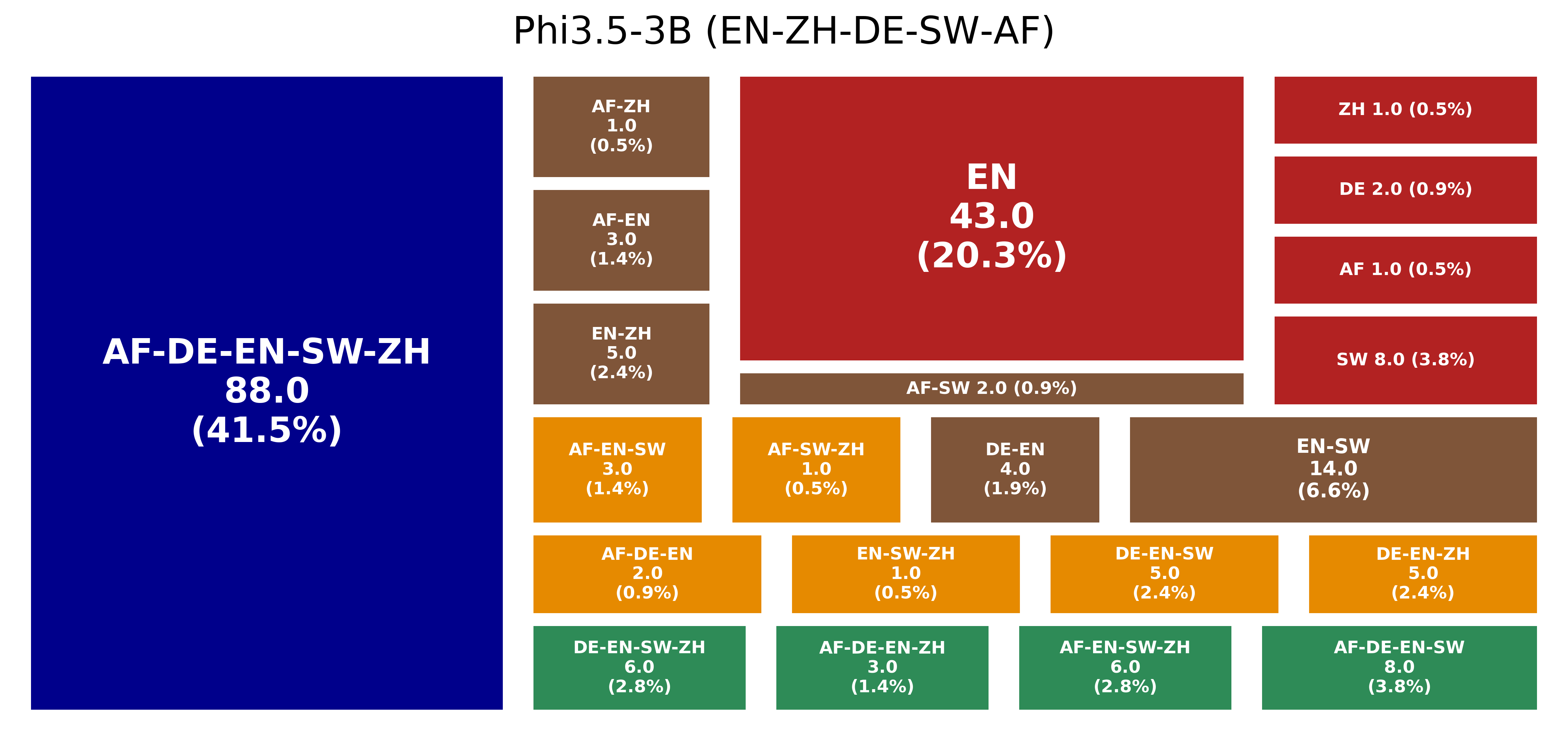}\hfill
\includegraphics[width=0.47\linewidth]{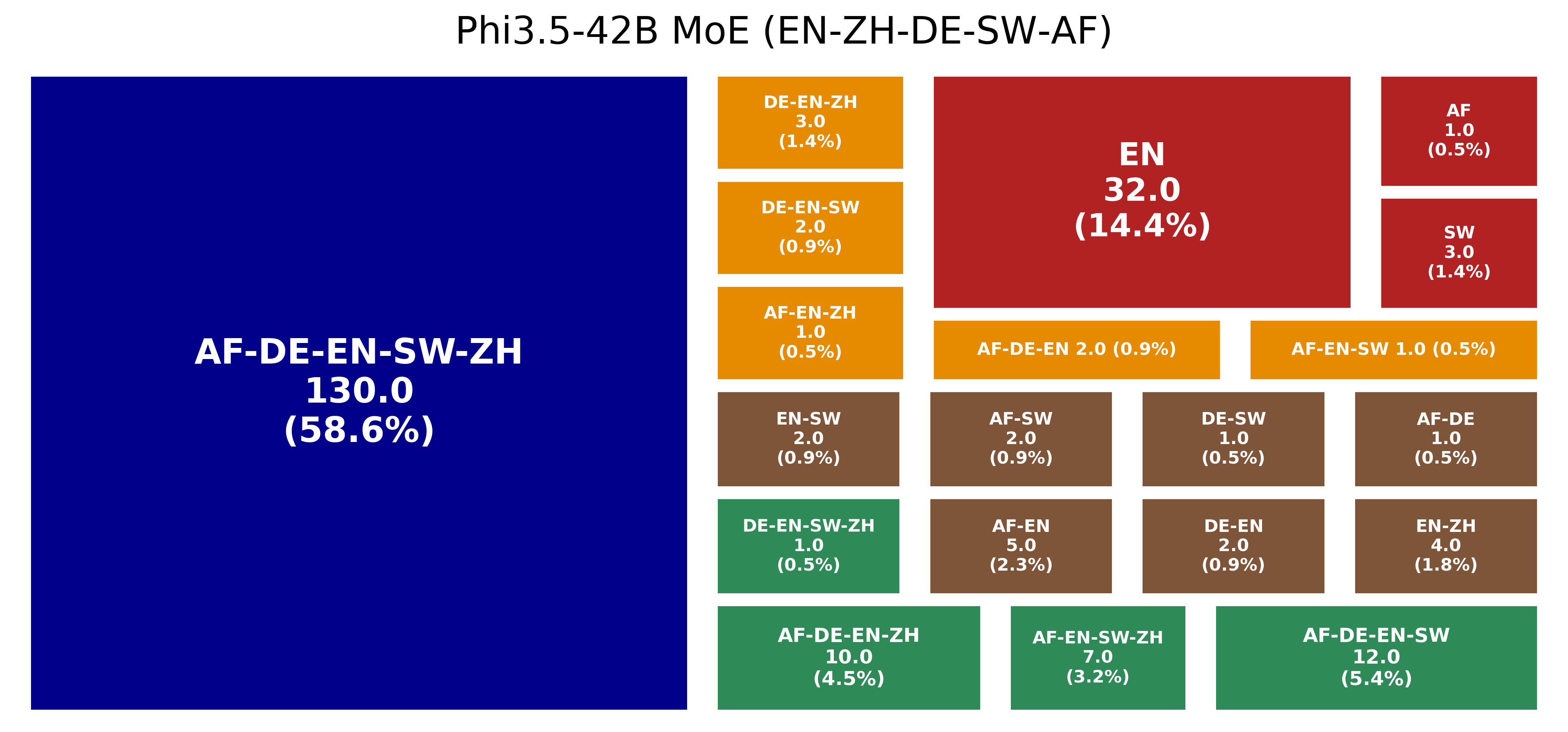}
\caption{To compare architectures, we study Phi3.5-3B MiniInstruct and Phi3.5-42B MoE (Mixture-of-Experts). For Phi3.5-42B MoE architecture, we observe that the proportions of shared attention heads are higher compared to language-exclusive attention heads.}
\label{fig:phi_moe}
\end{figure}

\begin{figure}[H]
\centering
\includegraphics[width=0.47\linewidth]{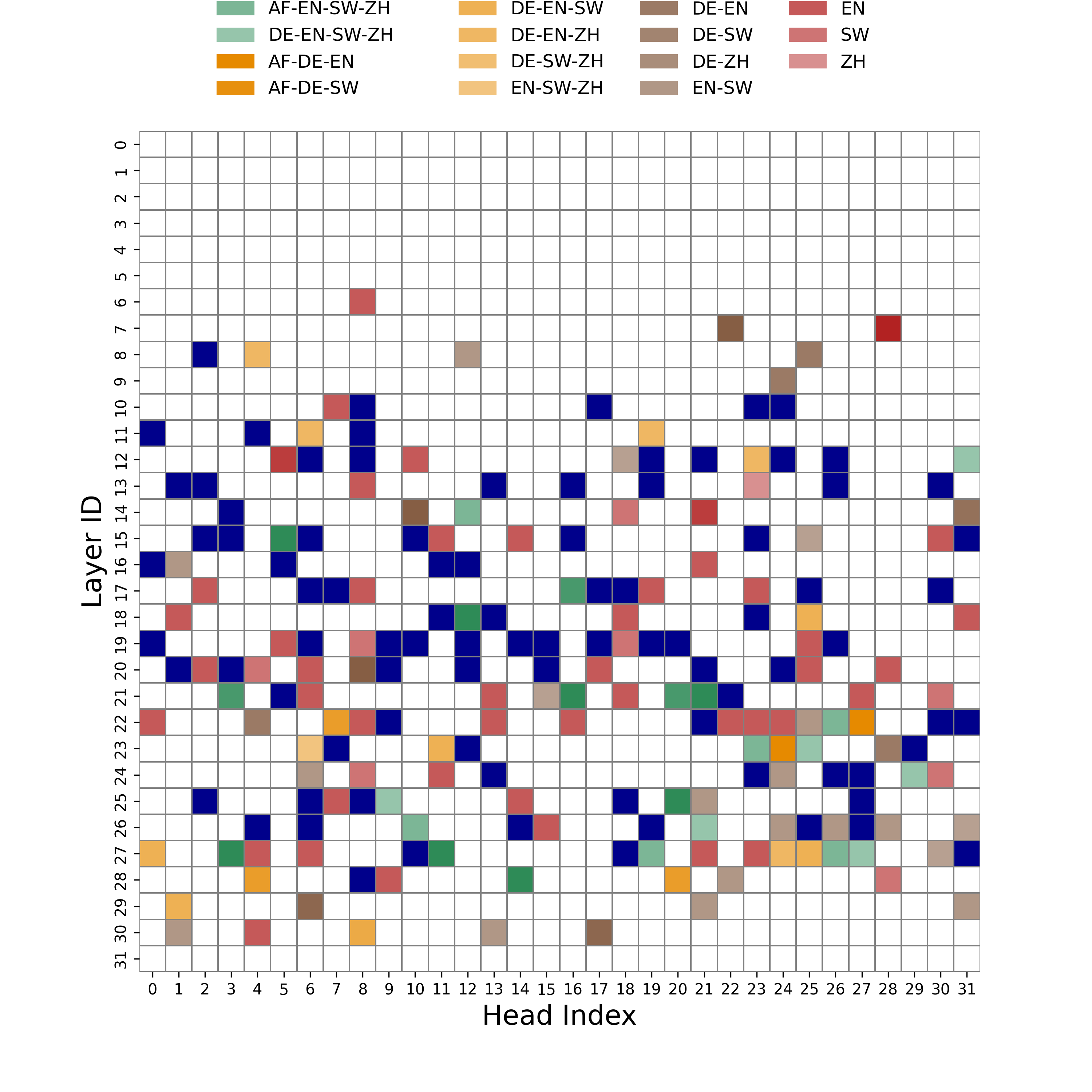}\hfill
\includegraphics[width=0.47\linewidth]{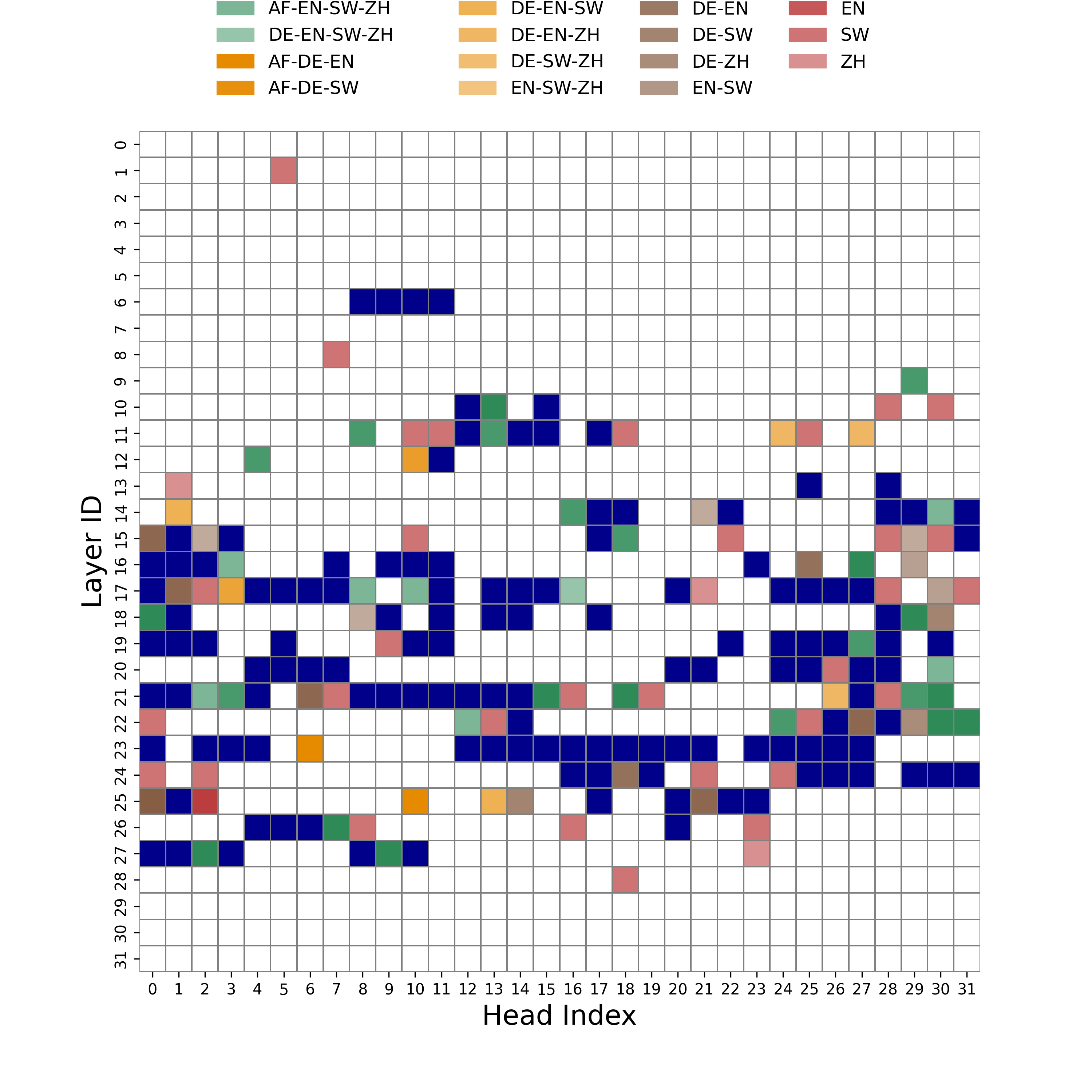}
\caption{To assess how scaling affects the placement of shared and language-exclusive attention heads, Phi3.5-3B MiniInstruct and Phi3.5--42B MoE (Mixture-of-Experts). The spatial distribution of these head types remains largely consistent across scales.}
\label{fig:phi_moe_location}
\end{figure}

\subsection{Considering the effect of including more low-resource languages}\label{appendix:more_lowres}

\begin{figure}[h!]
\centering
\includegraphics[width=0.75\linewidth]{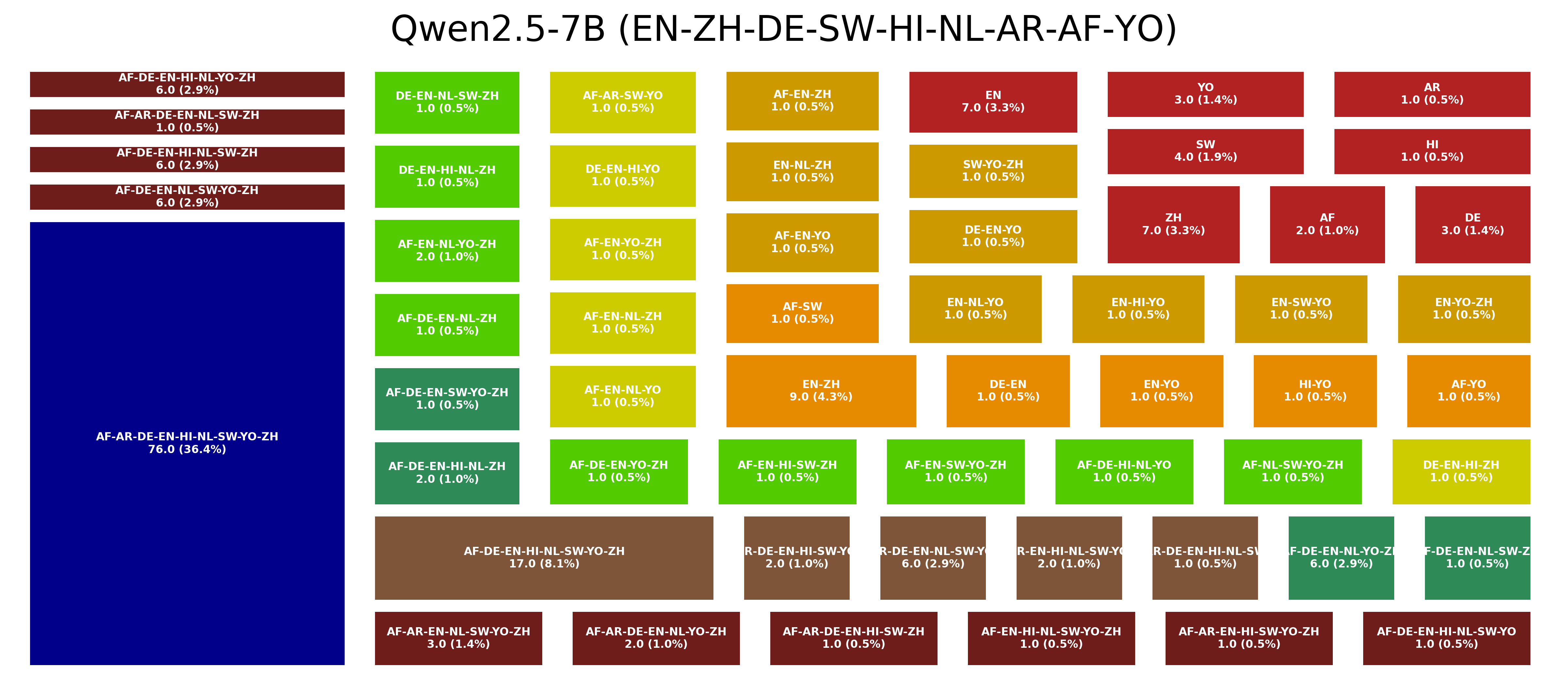}
\caption{To study the effect of adding more low-resource languages, we evaluate Qwen2.5-7B Instruct on English, Chinese, German, Swahili, Hindi, Dutch, and Arabic. As the language set expands, the fraction of shared attention heads decreases, while language-exclusive heads become more prevalent.}
\label{appendix-figs:qwen_7lang}
\end{figure}
\newpage

\section{Language Specificity and Training Data Composition}
\label{appendix:model_data}

The variance in retrieval head intersection observed in our analysis—ranging from the high degree of cross-lingual sharing in Qwen2.5-7B to the English-dominant specialization in Phi3.5-3B — can be interpreted through the lens of the distinct data-scaling strategies and training recipes employed by each model family. To provide context for the variability in retrieval head behavior, we summarize the known multilingual and pre-training configurations for the three models in Table~\ref{tab:model_comparison}.

\begin{table}[h!]
\centering
\caption{Comparison of training data scale and multilingual distribution. Data compiled from official model cards and technical reports (\citet{qwen2025qwen25technicalreport, abdin2024phi3technicalreporthighly, grattafiori2024Llama}).}
\label{tab:model_comparison}
\begin{small}
\begin{tabular}{@{}lccc@{}}
\toprule
\textbf{Feature} & \textbf{Qwen2.5 7B} & \textbf{Llama3.1 8B} & \textbf{Phi-3.5 3B Mini} \\ \midrule
Total Pre-training Tokens & 18.0 Trillion & 15.0 Trillion & 3.4 Trillion \\
Primary Language Focus & Balanced (EN/CN) & English-Centric & English Reasoning \\
Non-English Proportion & High ($\sim$40--50\% Est.) & $\sim$5\% ($\sim$750B tokens) & $\sim$10\% (Multilingual) \\
Post-training Strategy & Instruction/Reasoning & 3.01\% Multilingual SFT & Synthetic Textbooks \\
Multilingual Scope & 29+ Official Languages & 8+ Official (100+ total) & 20+ Languages \\ \bottomrule
\label{appendix-tab:model_train_configs}
\end{tabular}
\end{small}
\end{table}

We observe that the degree of cross-lingual head intersection is strongly correlated with pre-training data composition and scale. In Qwen2.5 7B, the massive 18T token corpus and high-density non-English data promote "universal" retrieval mechanisms, with 57.4\% of heads shared across all five studied languages. Conversely, Phi-3.5 3B exhibits greater linguistic specialization, with the lowest 5-language intersection (41.5\%) and a bias toward English-specific heads; this likely reflects its reliance on high-quality, primarily English synthetic data. These findings suggest that while scale drives language-agnostic representations, the specific data mixture determines whether a model develops universal or language-specific retrieval pathways.

\newpage
\section{Characteristics of Retrieval Transition Heads}\label{appendix:secD_1}
\subsection{Are Retrieval Transition heads universally present in Multilingual Language Models?} 
\begin{figure*}[h!]
\includegraphics[width=1.0\textwidth]{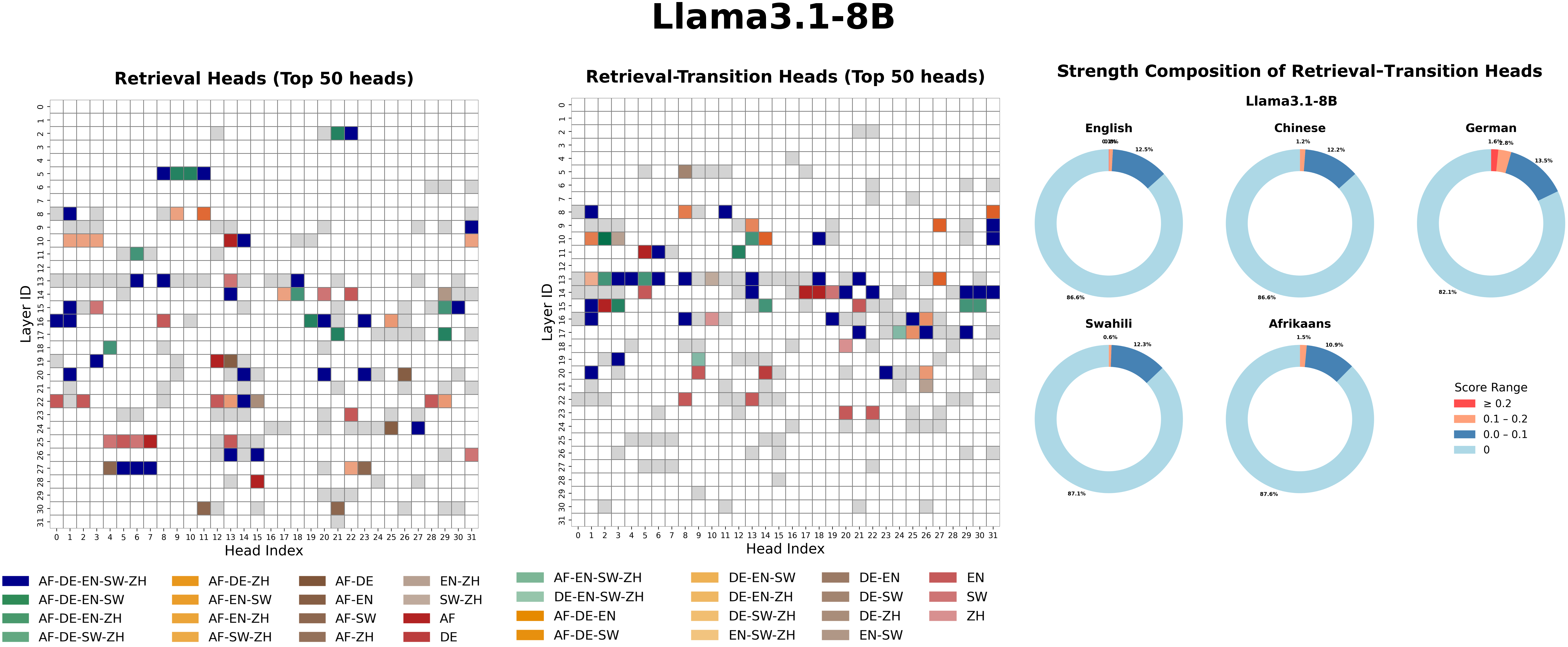}
\caption{Layer–head distributions of Retrieval Heads (RH) and Retrieval-Transition Heads (RTH) across languages for Llama3.1 8B Instruct. Colored cells are part of Top-50 most prominent heads ranked by score. Among these colored heads, we find language-specific RH are dominant in the final layers while RTH are prominent in the middle layers. Like RH heads, RTH heads too show \textit{sparsity} with only 0-5\% of heads with an RTS score above 0.1}
\label{fig:loc_rh_rth_Llama}
\end{figure*}

\begin{figure*}[h!]
\includegraphics[width=1.0\textwidth]{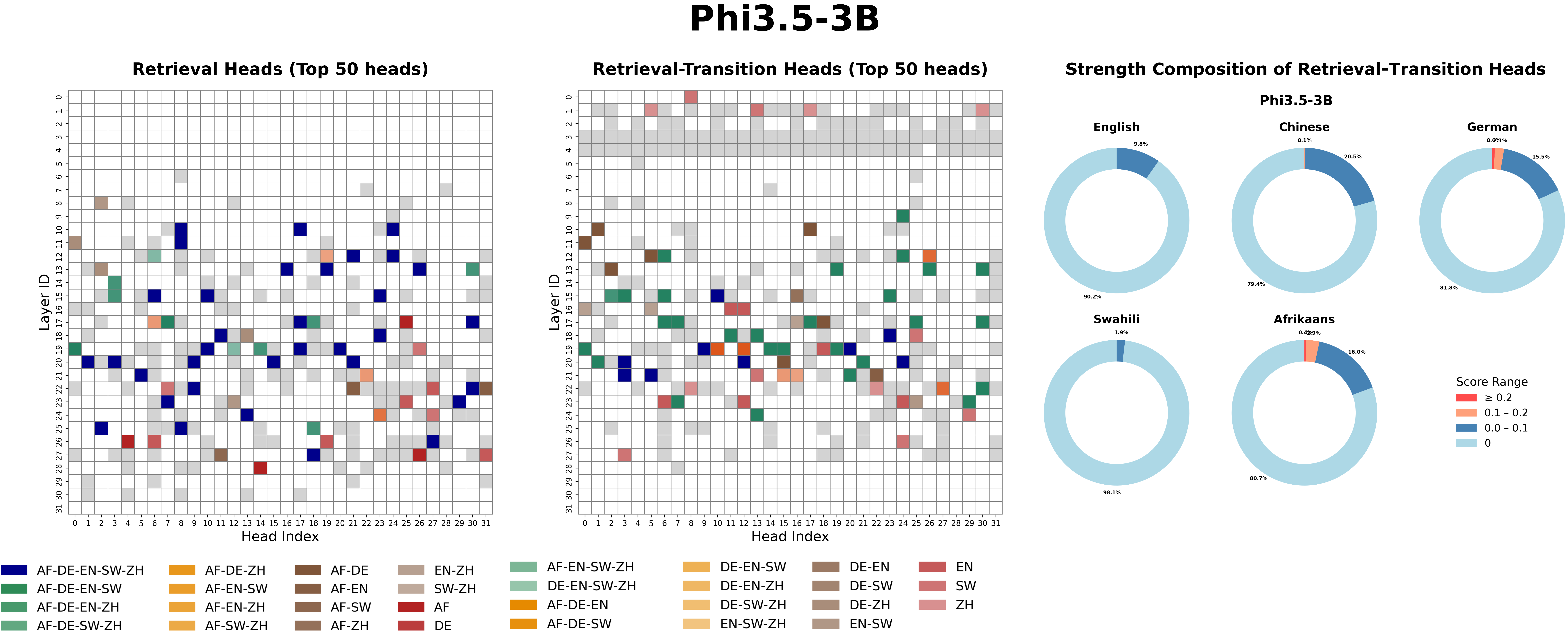}
\caption{Layer–head distributions of Retrieval Heads (RH) and Retrieval-Transition Heads (RTH) across languages for Phi-3.5 3B MiniInstruct. Colored cells are part of Top-50 most prominent heads ranked by score. Among these colored heads, we find language-specific RH are dominant in the final layers while RTH are prominent in the middle layers. Like RH heads, RTH heads too show \textit{sparsity} with only 0-3\% of heads with an RTS score above 0.1}
\label{fig:loc_rh_rth_phi}
\end{figure*}

\subsection{Effect of larger model size and MOE architecture on Retrieval Transition head distribution}

\begin{figure*}[h!]
\centering
\includegraphics[width=0.7\textwidth]{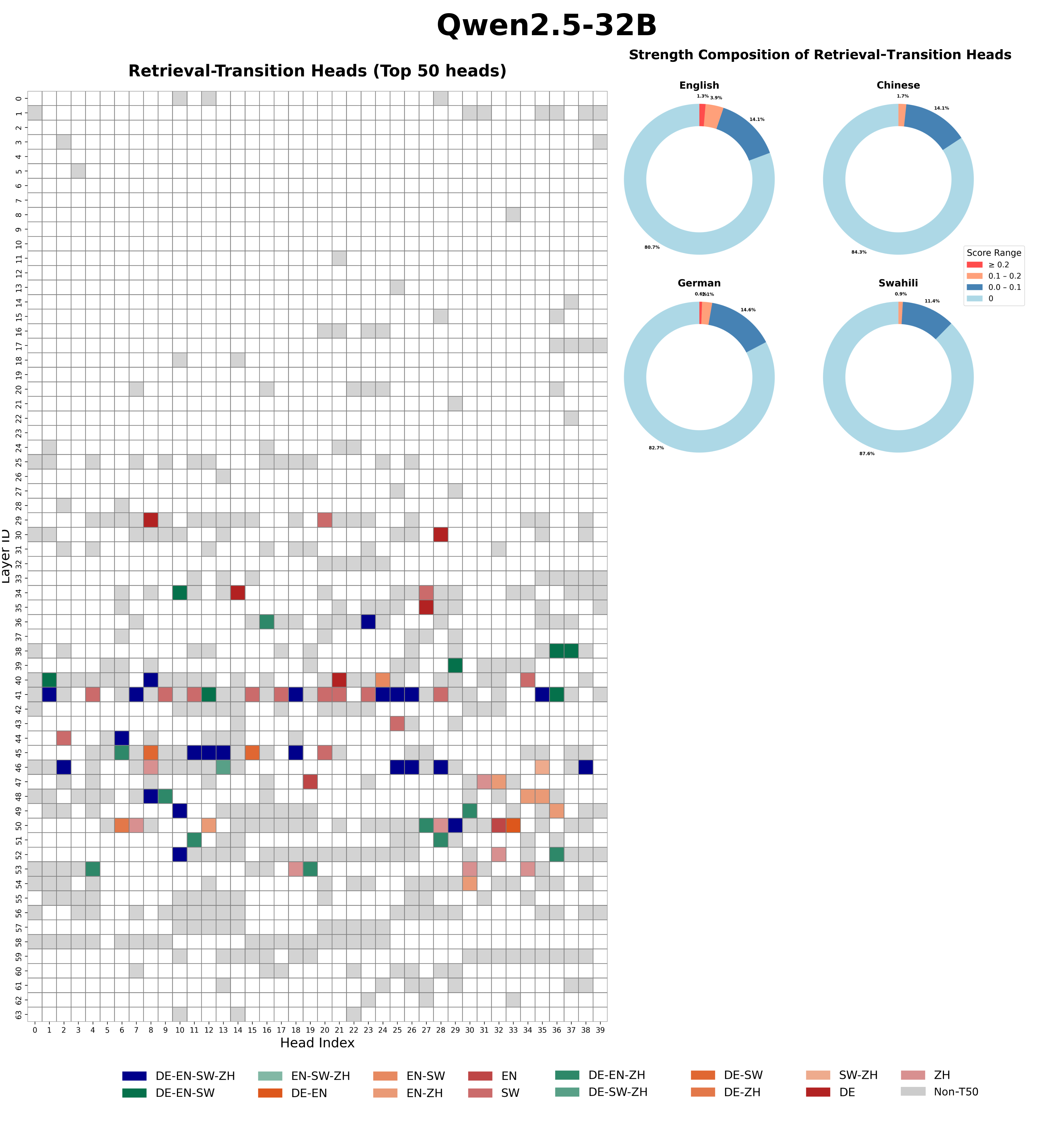}
\caption{Layer–head distributions of Retrieval-Transition Heads (RTH) across languages for Qwen-2.5 32B Instruct. Colored cells are part of Top-50 most prominent heads ranked by score. We find that RTH heads consistently appear in larger model sizes as well.}
\label{fig:loc_rh_rth_phi}
\end{figure*}

\begin{figure*}[h!]
\centering
\includegraphics[width=0.75\textwidth]{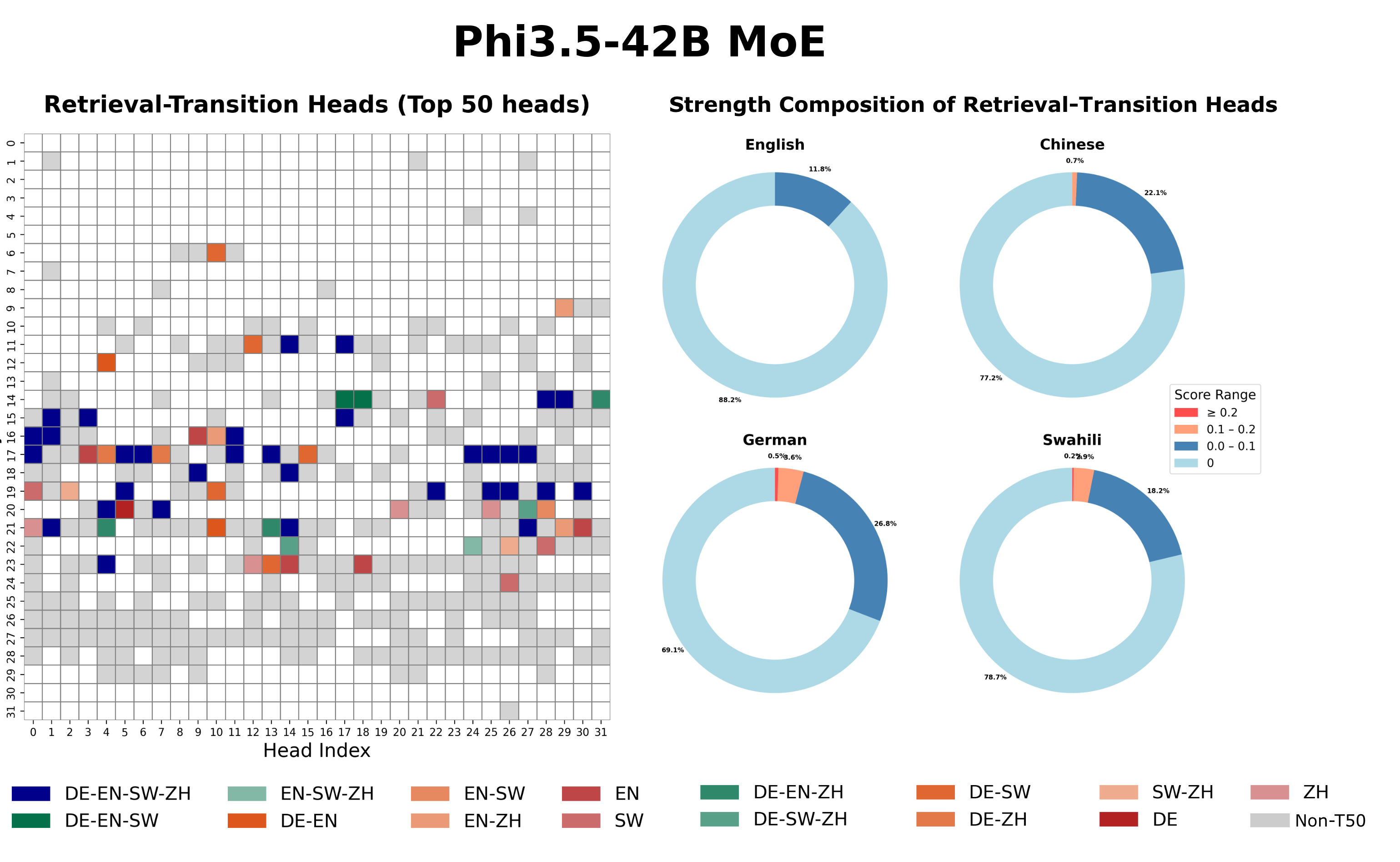}
\caption{Layer–head distributions of Retrieval-Transition Heads (RTH) across languages for Phi-3.5 42B MOE Instruct. Colored cells are part of Top-50 most prominent heads ranked by score. We find that RTH heads also appear in different model architectures like Mixture-of-Experts.}
\label{fig:loc_rh_rth_phi}
\end{figure*}

\newpage

\subsection{Do Retrieval Transition heads also appear in pretrained-only models?}

\begin{figure*}[h!]
\centering
\includegraphics[width=0.75\textwidth]{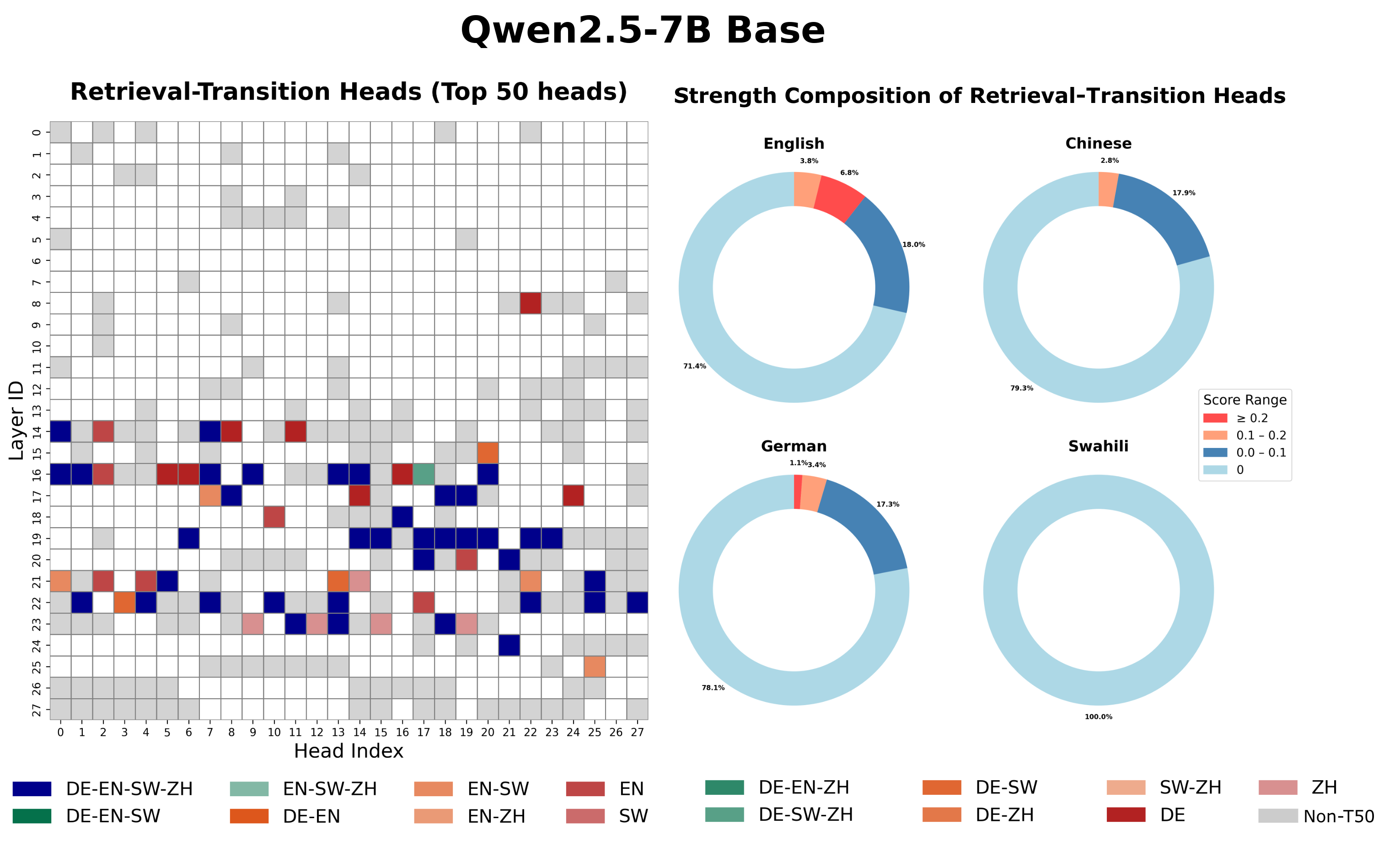}
\caption{Layer–head distributions of Retrieval-Transition Heads (RTH) across languages for Qwen2.5 7B Base. Colored cells are part of Top-50 most prominent heads ranked by score. We find that RTH heads are also present in non-Instruct models, however are primarily English language dominated.}
\label{fig:loc_rh_rth_phi}
\end{figure*}

\begin{figure*}[h!]
\centering
\includegraphics[width=0.75\textwidth]{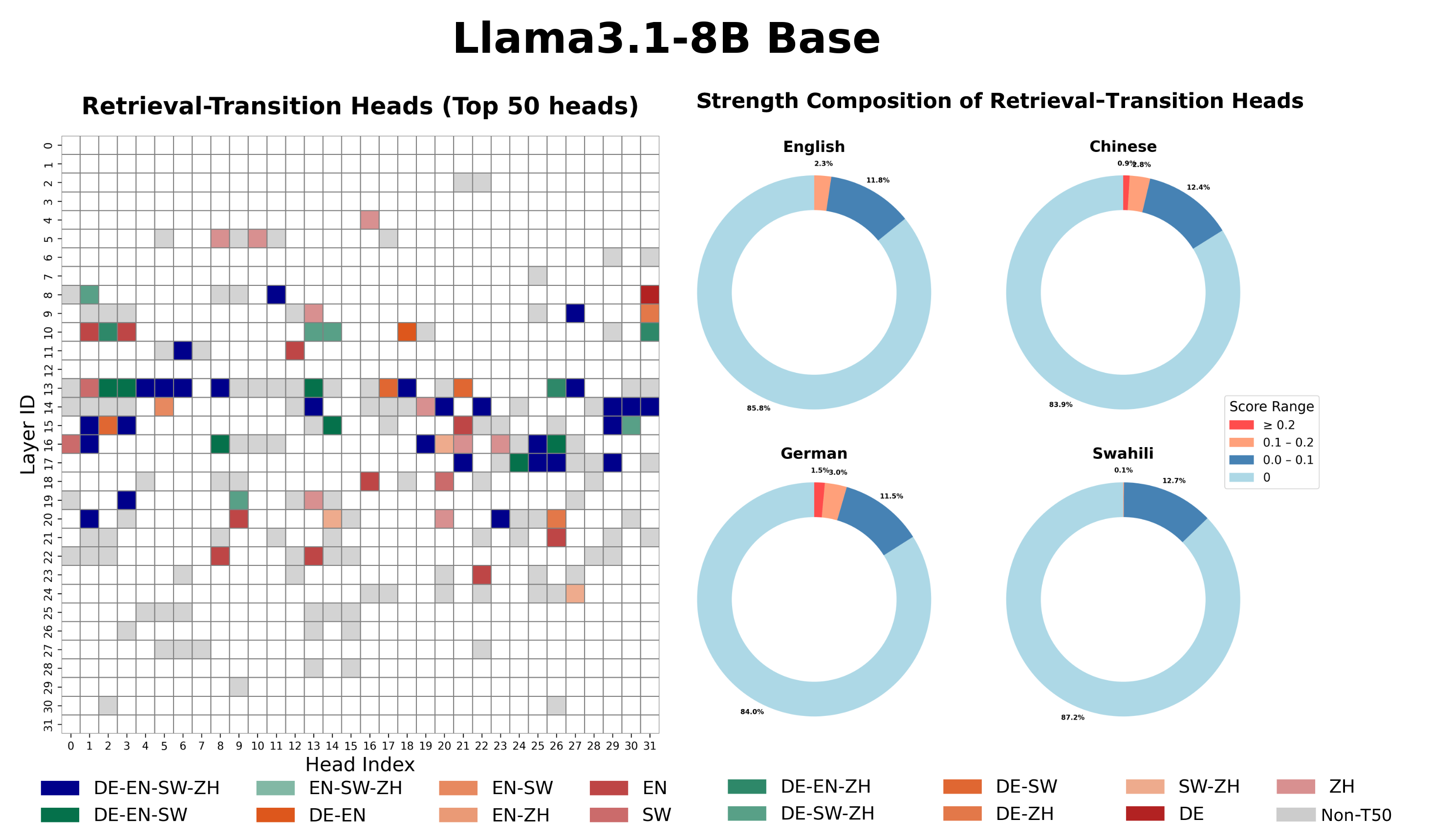}
\caption{Layer–head distributions of Retrieval-Transition Heads (RTH) across languages for Llama3.1 8B Base. Colored cells are part of Top-50 most prominent heads ranked by score. We find that RTH heads are also present in non-Instruct models, however are primarily English language dominated.}
\label{fig:loc_rh_rth_phi}
\end{figure*}

\subsection{Analyzing variability introduced by LLM-based aligner}\label{appendix:secD_2}


\begin{table}[h!]
\centering
\small
\begin{tabular}{lcccc}
\toprule
Language & Top-15 & Top-25 & Top-50 & Top-100 \\
\midrule
$RTH_{en}$ & 15 & 25  & 50 & 100 \\
$RTH_{zh}$ & 13  & 23 & 48 & 98 \\
\bottomrule
\end{tabular}
\caption{Intersection size of the top-$k$ retrieval-transition heads identified using \texttt{Qwen3-30B-A3B-Thinking-2507} versus \texttt{GPT-4.1} as the LLM-based aligner. We run the analysis on \texttt{Qwen-2.5-7B Instruct} for $RTH_{en}$ and $RTH_{zh}$.  The top-$k$ RTH sets largely overlap across aligners, indicating that head identification is robust to the choice of aligner.}

\label{tab:gpt_41_aligner}
\end{table}

\section{Human Validation of the Cross-Lingual Aligner} \label{app:aligner-validation}

\camera{We validate the aligner's output using a human judge for correctness, complementing comparison in Appendix~\ref{appendix:secD_2}.}

\paragraph{Setup.} \camera{We sample $50$ en$\to$zh examples uniformly at random from the population the aligner processed when computing $RTS$. We generate the mapping by running the same alignment function used in our pipeline.}

\paragraph{Annotation protocol.} \camera{An annotator fluent in English and Chinese judged every target subword in reading order, marking the aligner's links either correct, or incorrect together with the source subword indices it should have produced. The annotator also marked subwords having no source counterpart. Annotating every subword, rather than ticking each proposed link, is what makes recall measurable, because a link the aligner omits leaves no row to tick.}

\paragraph{Evaluation and Results.} \camera{As we assign credit for each decoding step, we report step-level scores. The precision, recall and F1 score is 0.67, 0.87 and 0.76 respectively. One of the dominating errors is under-abstention: the model's own commentary i.e. text that is not a translation of the needle is linked back into the needle rather than assigned null. Where the aligner abstains it is reliable, agreeing with the annotator $95.3\%$ of the time. Most are punctuation, copulas, and byte-level fragments from splitting Chinese characters, which alginer leaves unaligned rather than guessing. 
}

\rebuttal{\section{Causal Intervention through Attention Head Masking}\label{appendix:causal_method}}
\rebuttal{Following \cite{wu2024retrieval}, we quantify the causal impact of individual attention heads by performing targeted interventions during inference. Specifically, we perturb the attention logits of a predefined subset of heads prior to the softmax operation, effectively neutralizing their contribution to the subsequent layer representations.}

\rebuttal{Let $A^{(l,h)} \in \mathbb{R}^{q \times k}$ denote the pre-softmax attention logits for head $h$ in layer $l$, computed as:
\begin{equation}
A^{(l,h)} = \frac{Q^{(l,h)} K^{(l,h)\top}}{\sqrt{d_h}}
\end{equation}
where $Q^{(l,h)}$ and $K^{(l,h)}$ are the query and key projections, respectively. For a head designated for ablation, we replace its logit matrix with a large negative constant:
\begin{equation}
A^{(l,h)} \leftarrow -C \cdot \mathbf{1}_{q \times k}
\end{equation}}
\rebuttal{where $C$ is a sufficiently large scalar (we set $C = 10^{9}$). This intervention ensures that, upon softmax normalization, the resulting attention probabilities approach zero, effectively preventing the head from aggregating information from any key positions.}

\rebuttal{This masking is localized; all other heads and model parameters remain unmodified, and the residual stream is only affected by the absence of the ablated head's output. Operationally, we specify a set of $(\text{layer}, \text{head})$ tuples at inference time to selectively ablate different functional categories, such as retrieval or retrieval-transition heads, alongside randomly sampled heads as a control condition.}

\rebuttal{We evaluate the causal importance of these components by measuring the degradation in downstream benchmark performance relative to the original, unperturbed model. Significant performance loss under a specific mask is interpreted as evidence of that head's functional necessity for the task.}

\newpage

\section{Additional experiments and Qualitative analysis of Retrieval-Transition Heads}
\subsection{Effect of hyperparameter \textit{k} (number of heads masked)}\label{appendix:secE_1}

\begin{table*}[h!]
    \centering
    \definecolor{negcolor}{rgb}{0.70,0.13,0.13}
    \definecolor{poscolor}{rgb}{0.0,0.6,0.3}
    \definecolor{lightblue}{rgb}{0.9, 0.95, 1.0}

    \renewcommand{\arraystretch}{1.3}
    \footnotesize
    \setlength{\tabcolsep}{4pt}

    \resizebox{\textwidth}{!}{
    \begin{tabular}{c|l|cccccc|cccc}
    \toprule
    \qquad & \multirow{2}{*}{\normalsize{\textbf{Heads}}} & \multicolumn{6}{c|}{\normalsize{\textbf{MMLU-ProX (Acc.)}}} & \multicolumn{4}{c}{\normalsize{\textbf{MLQA (F1-score)}}} \\
    & & {\texttt{en}} & {\texttt{zh}} & {\texttt{de}} & {\texttt{sw}} & {\texttt{af}}  & \textbf{Avg} & {\texttt{en}} & {\texttt{zh}} & {\texttt{de}} & \textbf{Avg} \\
    \midrule

    \rowcolor{lightgray}\cellcolor{white}\multirow{4}{*}{\rotatebox{90}{\textbf{Qwen2.5 7B}}}
    & None
    & 42.56 & 44.22 & 40.57 & 22.59 & 32.98 & 36.58
    & 17.26 & 18.93 & 16.16 & 17.45 \\

    & Random
    & 42.85$^{\color{poscolor}+0.29}$ & 42.08$^{\color{negcolor}-2.14}$ & 29.94$^{\color{negcolor}-10.63}$ & 21.90$^{\color{negcolor}-0.69}$ & 29.65$^{\color{negcolor}-3.33}$ & 33.28$^{\color{negcolor}-3.30}$
    & 16.71$^{\color{negcolor}-0.55}$ & 17.87$^{\color{negcolor}-1.06}$ & 14.26$^{\color{negcolor}-1.90}$ & 16.28$^{\color{negcolor}-1.17}$ \\

    & RH
    & 39.85$^{\color{negcolor}-2.71}$ & 39.87$^{\color{negcolor}-4.35}$ & 37.08$^{\color{negcolor}-3.49}$ & 20.09$^{\color{negcolor}-2.50}$ & 29.82$^{\color{negcolor}-3.16}$ & 33.34$^{\color{negcolor}-3.24}$
    & 15.82$^{\color{negcolor}-1.44}$ & 18.28$^{\color{negcolor}-0.65}$ & 15.11$^{\color{negcolor}-1.05}$ & 16.40$^{\color{negcolor}-1.05}$ \\

    & \cellcolor{lightblue}RTH
    & \cellcolor{lightblue}30.64$^{\color{negcolor}-11.92}$ & \cellcolor{lightblue}35.37$^{\color{negcolor}-8.85}$ & \cellcolor{lightblue}24.83$^{\color{negcolor}-15.74}$ & \cellcolor{lightblue}15.58$^{\color{negcolor}-7.01}$ & \cellcolor{lightblue}19.02$^{\color{negcolor}-13.96}$ & \cellcolor{lightblue}\textbf{25.09}$^{\color{negcolor}\textbf{-11.50}}$
    & \cellcolor{lightblue}13.69$^{\color{negcolor}-3.57}$ & \cellcolor{lightblue}15.17$^{\color{negcolor}-3.76}$ & \cellcolor{lightblue}8.95$^{\color{negcolor}-7.21}$ & \cellcolor{lightblue}\textbf{12.60}$^{\color{negcolor}\textbf{-4.85}}$ \\

    \midrule

    \rowcolor{lightgray}\cellcolor{white}\multirow{4}{*}{\rotatebox{90}{\textbf{Llama3.1 8B}}}
    & None
    & 43.59 & 31.41 & 30.89 & 13.27 & 27.38 & 29.31
    & 39.05 & 49.49 & 36.76 & 41.77 \\

    & Random
    & 43.17$^{\color{negcolor}-0.42}$ & 31.47$^{\color{poscolor}+0.06}$ & 26.94$^{\color{negcolor}-3.95}$ & 14.60$^{\color{poscolor}+1.33}$ & 20.89$^{\color{negcolor}-6.49}$ & 27.41$^{\color{negcolor}-1.89}$
    & 40.35$^{\color{poscolor}+1.30}$ & 50.11$^{\color{poscolor}+0.62}$ & 39.20$^{\color{poscolor}+2.44}$ & 43.22$^{\color{poscolor}+1.45}$ \\

    & RH
    & 40.23$^{\color{negcolor}-3.36}$ & 26.97$^{\color{negcolor}-4.44}$ & 22.42$^{\color{negcolor}-8.47}$ & 1.70$^{\color{negcolor}-11.57}$ & 27.36$^{\color{negcolor}-0.02}$ & 23.74$^{\color{negcolor}-5.57}$
    & 39.08$^{\color{poscolor}+0.03}$ & 46.82$^{\color{negcolor}-2.67}$ & 36.33$^{\color{negcolor}-0.43}$ & 40.74$^{\color{negcolor}-1.02}$ \\

    & \cellcolor{lightblue}RTH
    & \cellcolor{lightblue}9.57$^{\color{negcolor}-34.02}$ & \cellcolor{lightblue}26.27$^{\color{negcolor}-5.14}$ & \cellcolor{lightblue}3.48$^{\color{negcolor}-27.41}$ & \cellcolor{lightblue}4.13$^{\color{negcolor}-9.14}$ & \cellcolor{lightblue}0.94$^{\color{negcolor}-26.44}$ & \cellcolor{lightblue}\textbf{8.88}$^{\color{negcolor}\textbf{-20.43}}$
    & \cellcolor{lightblue}30.88$^{\color{negcolor}-8.17}$ & \cellcolor{lightblue}43.83$^{\color{negcolor}-5.66}$ & \cellcolor{lightblue}36.02$^{\color{negcolor}-0.74}$ & \cellcolor{lightblue}\textbf{36.91}$^{\color{negcolor}\textbf{-4.86}}$ \\

    \bottomrule
    \end{tabular}
    }

    \vspace{6pt}

    \resizebox{\textwidth}{!}{%
    \begin{tabular}{c|l|ccccc|cccc}
    \toprule
    \qquad & \multirow{2}{*}{\normalsize{\textbf{Heads}}} & \multicolumn{5}{c|}{\normalsize{\textbf{MGSM (Acc.)}}} & \multicolumn{4}{c}{\normalsize{\textbf{XQuAD (F1-score)}}} \\
    & & {\texttt{en}} & {\texttt{zh}} & {\texttt{de}} & {\texttt{sw}} & \textbf{Avg} & {\texttt{en}} & {\texttt{zh}} & {\texttt{de}} & \textbf{Avg} \\
    \midrule

    \rowcolor{lightgray}\cellcolor{white}\multirow{4}{*}{\rotatebox{90}{\textbf{Qwen2.5 7B}}}
    & None
    & 33.6 & 31.6 & 27.2 & 6.4 & 24.70
    & 15.72 & 0.82 & 14.28 & 10.27 \\

    & Random
    & 29.6$^{\color{negcolor}-4.0}$ & 26.8$^{\color{negcolor}-4.8}$ & 18.0$^{\color{negcolor}-9.2}$ & 2.4$^{\color{negcolor}-4.0}$ & 19.20$^{\color{negcolor}-5.50}$
    & 13.82$^{\color{negcolor}-1.90}$ & 0.88$^{\color{poscolor}+0.06}$ & 10.85$^{\color{negcolor}-3.43}$ & 8.52$^{\color{negcolor}-1.76}$ \\

    & RH
    & 36.4$^{\color{poscolor}+2.8}$ & 22.4$^{\color{negcolor}-9.2}$ & 23.2$^{\color{negcolor}-4.0}$ & 3.6$^{\color{negcolor}-2.8}$ & 21.40$^{\color{negcolor}-3.30}$
    & 13.89$^{\color{negcolor}-1.83}$ & 0.56$^{\color{negcolor}-0.26}$ & 12.64$^{\color{negcolor}-1.64}$ & 9.03$^{\color{negcolor}-1.24}$ \\

    & \cellcolor{lightblue}RTH
    & \cellcolor{lightblue}18.0$^{\color{negcolor}-15.6}$ & \cellcolor{lightblue}23.6$^{\color{negcolor}-8.0}$ & \cellcolor{lightblue}5.6$^{\color{negcolor}-21.6}$ & \cellcolor{lightblue}2.0$^{\color{negcolor}-4.4}$ & \cellcolor{lightblue}\textbf{12.30}$^{\color{negcolor}\textbf{-12.40}}$
    & \cellcolor{lightblue}12.04$^{\color{negcolor}-3.68}$ & \cellcolor{lightblue}0.46$^{\color{negcolor}-0.36}$ & \cellcolor{lightblue}7.34$^{\color{negcolor}-6.94}$ & \cellcolor{lightblue}\textbf{6.61}$^{\color{negcolor}\textbf{-3.66}}$ \\

    \midrule

    \rowcolor{lightgray}\cellcolor{white}\multirow{4}{*}{\rotatebox{90}{\textbf{Llama3.1 8B}}}
    & None
    & 73.2 & 62.4 & 59.2 & 48.0 & 60.70
    & 35.26 & 28.70 & 31.10 & 31.69 \\

    & Random
    & 72.4$^{\color{negcolor}-0.8}$ & 58.4$^{\color{negcolor}-4.0}$ & 57.6$^{\color{negcolor}-1.6}$ & 28.4$^{\color{negcolor}-19.6}$ & 54.20$^{\color{negcolor}-6.50}$
    & 37.28$^{\color{poscolor}+2.02}$ & 28.81$^{\color{poscolor}+0.11}$ & 34.46$^{\color{poscolor}+3.36}$ & 33.52$^{\color{poscolor}+1.83}$ \\

    & RH
    & 72.0$^{\color{negcolor}-1.2}$ & 62.4$^{\phantom{+}0.0}$ & 52.4$^{\color{negcolor}-6.8}$ & 48.0$^{\phantom{+}0.0}$ & 58.70$^{\color{negcolor}-2.00}$
    & 36.68$^{\color{poscolor}+1.42}$ & 21.43$^{\color{negcolor}-7.27}$ & 29.52$^{\color{negcolor}-1.58}$ & \textbf{29.21}$^{\color{negcolor}\textbf{-2.48}}$ \\

    & \cellcolor{lightblue}RTH
    & \cellcolor{lightblue}21.2$^{\color{negcolor}-52.0}$ & \cellcolor{lightblue}24.8$^{\color{negcolor}-37.6}$ & \cellcolor{lightblue}25.6$^{\color{negcolor}-33.6}$ & \cellcolor{lightblue}19.6$^{\color{negcolor}-28.4}$ & \cellcolor{lightblue}\textbf{22.80}$^{\color{negcolor}\textbf{-37.90}}$
    & \cellcolor{lightblue}28.48$^{\color{negcolor}-6.78}$ & \cellcolor{lightblue}28.70$^{\phantom{+}0.00}$ & \cellcolor{lightblue}30.88$^{\color{negcolor}-0.22}$ & \cellcolor{lightblue}29.35$^{\color{negcolor}-2.33}$ \\

    \bottomrule
    \end{tabular}
    }

    \caption{Impact of top-15 head masking on MMLU-ProX, MGSM, MLQA, and XQuAD using Qwen2.5 7B Instruct and Llama3.1 8B Instruct. We report accuracy for MMLU-ProX and MGSM, and F1 score for MLQA and XQuAD, computed using the lm-evaluation-harness. The \textbf{Avg} column reports the mean across the evaluated languages, with the colored superscript denoting the absolute change relative to the no-masking baseline.}
    \label{tab:masking_results_top15}
\end{table*}

\begin{table*}[h!]
    \centering
    \definecolor{negcolor}{rgb}{0.70,0.13,0.13}
    \definecolor{poscolor}{rgb}{0.0,0.6,0.3}
    \definecolor{lightblue}{rgb}{0.9, 0.95, 1.0}

    \renewcommand{\arraystretch}{1.3}
    \footnotesize
    \setlength{\tabcolsep}{4pt}

    \resizebox{\textwidth}{!}{
    \begin{tabular}{c|l|cccccc|cccc}
    \toprule
    \qquad & \multirow{2}{*}{\normalsize{\textbf{Heads}}} & \multicolumn{6}{c|}{\normalsize{\textbf{MMLU-ProX (Acc.)}}} & \multicolumn{4}{c}{\normalsize{\textbf{MLQA (F1-score)}}} \\
    & & {\texttt{en}} & {\texttt{zh}} & {\texttt{de}} & {\texttt{sw}} & {\texttt{af}} & \textbf{Avg} & {\texttt{en}} & {\texttt{zh}} & {\texttt{de}} & \textbf{Avg} \\
    \midrule

    \rowcolor{lightgray}\cellcolor{white}\multirow{4}{*}{\rotatebox{90}{\textbf{Qwen2.5 7B}}}
    & None
    & 42.56 & 44.22 & 40.57 & 22.59 & 32.98 & 36.58
    & 17.26 & 18.93 & 16.16 & 17.45 \\

    & Random
    & 40.95$^{\color{negcolor}-1.61}$ & 39.71$^{\color{negcolor}-4.51}$ & 35.75$^{\color{negcolor}-4.82}$ & 22.31$^{\color{negcolor}-0.28}$ & 29.84$^{\color{negcolor}-3.14}$ & 33.71$^{\color{negcolor}-2.87}$
    & 17.33$^{\color{poscolor}+0.07}$ & 18.16$^{\color{negcolor}-0.77}$ & 16.27$^{\color{poscolor}+0.11}$ & 17.25$^{\color{negcolor}-0.20}$ \\

    & RH
    & 19.48$^{\color{negcolor}-23.08}$ & 18.98$^{\color{negcolor}-25.24}$ & 14.84$^{\color{negcolor}-25.73}$ & 0.30$^{\color{negcolor}-22.29}$ & 8.10$^{\color{negcolor}-24.88}$ & 12.34$^{\color{negcolor}-24.24}$
    & 6.23$^{\color{negcolor}-11.03}$ & 6.63$^{\color{negcolor}-12.30}$ & 5.05$^{\color{negcolor}-11.11}$ & 5.97$^{\color{negcolor}-11.48}$ \\

    & \cellcolor{lightblue}RTH
    & \cellcolor{lightblue}14.14$^{\color{negcolor}-28.42}$ & \cellcolor{lightblue}24.13$^{\color{negcolor}-20.09}$ & \cellcolor{lightblue}3.24$^{\color{negcolor}-37.33}$ & \cellcolor{lightblue}11.05$^{\color{negcolor}-11.54}$ & \cellcolor{lightblue}5.46$^{\color{negcolor}-27.52}$ & \cellcolor{lightblue}\textbf{11.60}$^{\color{negcolor}\textbf{-24.98}}$
    & \cellcolor{lightblue}5.03$^{\color{negcolor}-12.23}$ & \cellcolor{lightblue}5.39$^{\color{negcolor}-13.54}$ & \cellcolor{lightblue}5.04$^{\color{negcolor}-11.12}$ & \cellcolor{lightblue}\textbf{5.15}$^{\color{negcolor}\textbf{-12.30}}$ \\

    \midrule

    \rowcolor{lightgray}\cellcolor{white}\multirow{4}{*}{\rotatebox{90}{\textbf{Llama3.1 8B}}}
    & None
    & 43.59 & 31.41 & 30.89 & 13.27 & 27.38 & 29.31
    & 39.05 & 49.49 & 36.76 & 41.77 \\

    & Random
    & 36.98$^{\color{negcolor}-6.61}$ & 27.28$^{\color{negcolor}-4.13}$ & 16.92$^{\color{negcolor}-13.97}$ & 5.60$^{\color{negcolor}-7.67}$ & 29.84$^{\color{poscolor}+2.46}$ & 23.32$^{\color{negcolor}-5.98}$
    & 38.53$^{\color{negcolor}-0.52}$ & 46.12$^{\color{negcolor}-3.37}$ & 37.93$^{\color{poscolor}+1.17}$ & 40.86$^{\color{negcolor}-0.91}$ \\

    & RH
    & 15.32$^{\color{negcolor}-28.27}$ & 10.83$^{\color{negcolor}-20.58}$ & 1.25$^{\color{negcolor}-29.64}$ & 0.80$^{\color{negcolor}-12.47}$ & 6.81$^{\color{negcolor}-20.57}$ & 7.00$^{\color{negcolor}-22.31}$
    & 31.13$^{\color{negcolor}-7.92}$ & 43.17$^{\color{negcolor}-6.32}$ & 22.88$^{\color{negcolor}-13.88}$ & 32.39$^{\color{negcolor}-9.37}$ \\

    & \cellcolor{lightblue}RTH
    & \cellcolor{lightblue}5.92$^{\color{negcolor}-37.67}$ & \cellcolor{lightblue}20.48$^{\color{negcolor}-10.93}$ & \cellcolor{lightblue}1.40$^{\color{negcolor}-29.49}$ & \cellcolor{lightblue}2.26$^{\color{negcolor}-11.01}$ & \cellcolor{lightblue}1.11$^{\color{negcolor}-26.27}$ & \cellcolor{lightblue}\textbf{6.23}$^{\color{negcolor}\textbf{-23.07}}$
    & \cellcolor{lightblue}22.12$^{\color{negcolor}-16.93}$ & \cellcolor{lightblue}33.51$^{\color{negcolor}-15.98}$ & \cellcolor{lightblue}26.56$^{\color{negcolor}-10.20}$ & \cellcolor{lightblue}\textbf{27.40}$^{\color{negcolor}\textbf{-14.37}}$ \\

    \bottomrule
    \end{tabular}
    }

    \vspace{6pt}

    \resizebox{\textwidth}{!}{%
    \begin{tabular}{c|l|ccccc|cccc}
    \toprule
    \qquad & \multirow{2}{*}{\normalsize{\textbf{Heads}}} & \multicolumn{5}{c|}{\normalsize{\textbf{MGSM (Acc.)}}} & \multicolumn{4}{c}{\normalsize{\textbf{XQuAD (F1-score)}}} \\
    & & {\texttt{en}} & {\texttt{zh}} & {\texttt{de}} & {\texttt{sw}} & \textbf{Avg} & {\texttt{en}} & {\texttt{zh}} & {\texttt{de}} & \textbf{Avg} \\
    \midrule

    \rowcolor{lightgray}\cellcolor{white}\multirow{4}{*}{\rotatebox{90}{\textbf{Qwen2.5 7B}}}
    & None
    & 33.6 & 31.6 & 27.2 & 6.4 & 24.70
    & 15.72 & 0.82 & 14.28 & 10.27 \\

    & Random
    & 16.4$^{\color{negcolor}-17.2}$ & 18.4$^{\color{negcolor}-13.2}$ & 13.2$^{\color{negcolor}-14.0}$ & 4.4$^{\color{negcolor}-2.0}$ & 13.10$^{\color{negcolor}-11.60}$
    & 14.96$^{\color{negcolor}-0.76}$ & 0.67$^{\color{negcolor}-0.15}$ & 12.67$^{\color{negcolor}-1.61}$ & 9.43$^{\color{negcolor}-0.84}$ \\

    & RH
    & 4.4$^{\color{negcolor}-29.2}$ & 5.6$^{\color{negcolor}-26.0}$ & 5.6$^{\color{negcolor}-21.6}$ & 1.2$^{\color{negcolor}-5.2}$ & 4.20$^{\color{negcolor}-20.50}$
    & 5.57$^{\color{negcolor}-10.15}$ & 0.41$^{\color{negcolor}-0.41}$ & 4.29$^{\color{negcolor}-9.99}$ & \textbf{3.42}$^{\color{negcolor}\textbf{-6.85}}$ \\

    & \cellcolor{lightblue}RTH
    & \cellcolor{lightblue}2.0$^{\color{negcolor}-31.6}$ & \cellcolor{lightblue}0.8$^{\color{negcolor}-30.8}$ & \cellcolor{lightblue}2.0$^{\color{negcolor}-25.2}$ & \cellcolor{lightblue}0.4$^{\color{negcolor}-6.0}$ & \cellcolor{lightblue}\textbf{1.30}$^{\color{negcolor}\textbf{-23.40}}$
    & \cellcolor{lightblue}4.86$^{\color{negcolor}-10.86}$ & \cellcolor{lightblue}0.34$^{\color{negcolor}-0.48}$ & \cellcolor{lightblue}5.19$^{\color{negcolor}-9.09}$ & \cellcolor{lightblue}3.46$^{\color{negcolor}-6.81}$ \\

    \midrule

    \rowcolor{lightgray}\cellcolor{white}\multirow{4}{*}{\rotatebox{90}{\textbf{Llama3.1 8B}}}
    & None
    & 73.2 & 62.4 & 59.2 & 48.0 & 60.70
    & 35.26 & 28.70 & 31.10 & 31.69 \\

    & Random
    & 67.2$^{\color{negcolor}-6.0}$ & 45.2$^{\color{negcolor}-17.2}$ & 41.6$^{\color{negcolor}-17.6}$ & 14.0$^{\color{negcolor}-34.0}$ & 42.00$^{\color{negcolor}-18.70}$
    & 36.47$^{\color{poscolor}+1.21}$ & 20.81$^{\color{negcolor}-7.89}$ & 33.30$^{\color{poscolor}+2.20}$ & 30.19$^{\color{negcolor}-1.49}$ \\

    & RH
    & 20.8$^{\color{negcolor}-52.4}$ & 25.6$^{\color{negcolor}-36.8}$ & 2.0$^{\color{negcolor}-57.2}$ & 4.0$^{\color{negcolor}-44.0}$ & 13.10$^{\color{negcolor}-47.60}$
    & 29.16$^{\color{negcolor}-6.10}$ & 22.49$^{\color{negcolor}-6.21}$ & 18.84$^{\color{negcolor}-12.26}$ & 23.50$^{\color{negcolor}-8.19}$ \\

    & \cellcolor{lightblue}RTH
    & \cellcolor{lightblue}2.8$^{\color{negcolor}-70.4}$ & \cellcolor{lightblue}1.6$^{\color{negcolor}-60.8}$ & \cellcolor{lightblue}2.4$^{\color{negcolor}-56.8}$ & \cellcolor{lightblue}3.2$^{\color{negcolor}-44.8}$ & \cellcolor{lightblue}\textbf{2.50}$^{\color{negcolor}\textbf{-58.20}}$
    & \cellcolor{lightblue}18.04$^{\color{negcolor}-17.22}$ & \cellcolor{lightblue}18.24$^{\color{negcolor}-10.46}$ & \cellcolor{lightblue}19.87$^{\color{negcolor}-11.23}$ & \cellcolor{lightblue}\textbf{18.72}$^{\color{negcolor}\textbf{-12.97}}$ \\

    \bottomrule
    \end{tabular}
    }

    \caption{Impact of top-50 head masking on MMLU-ProX, MGSM, MLQA, and XQuAD using Qwen2.5 7B Instruct and Llama3.1 8B Instruct. We report accuracy for MMLU-ProX and MGSM, and F1 score for MLQA and XQuAD, computed using the lm-evaluation-harness. The \textbf{Avg} column reports the mean across the evaluated languages, with the colored superscript denoting the absolute change relative to the no-masking baseline.}
    \label{tab:masking_results}
\end{table*}

\clearpage
\newpage

\subsection{LLM-as-a-judge prompt for failure-mode categorization}

\begin{figure}[h!]
    \centering
    \includegraphics[width=1.0\linewidth]{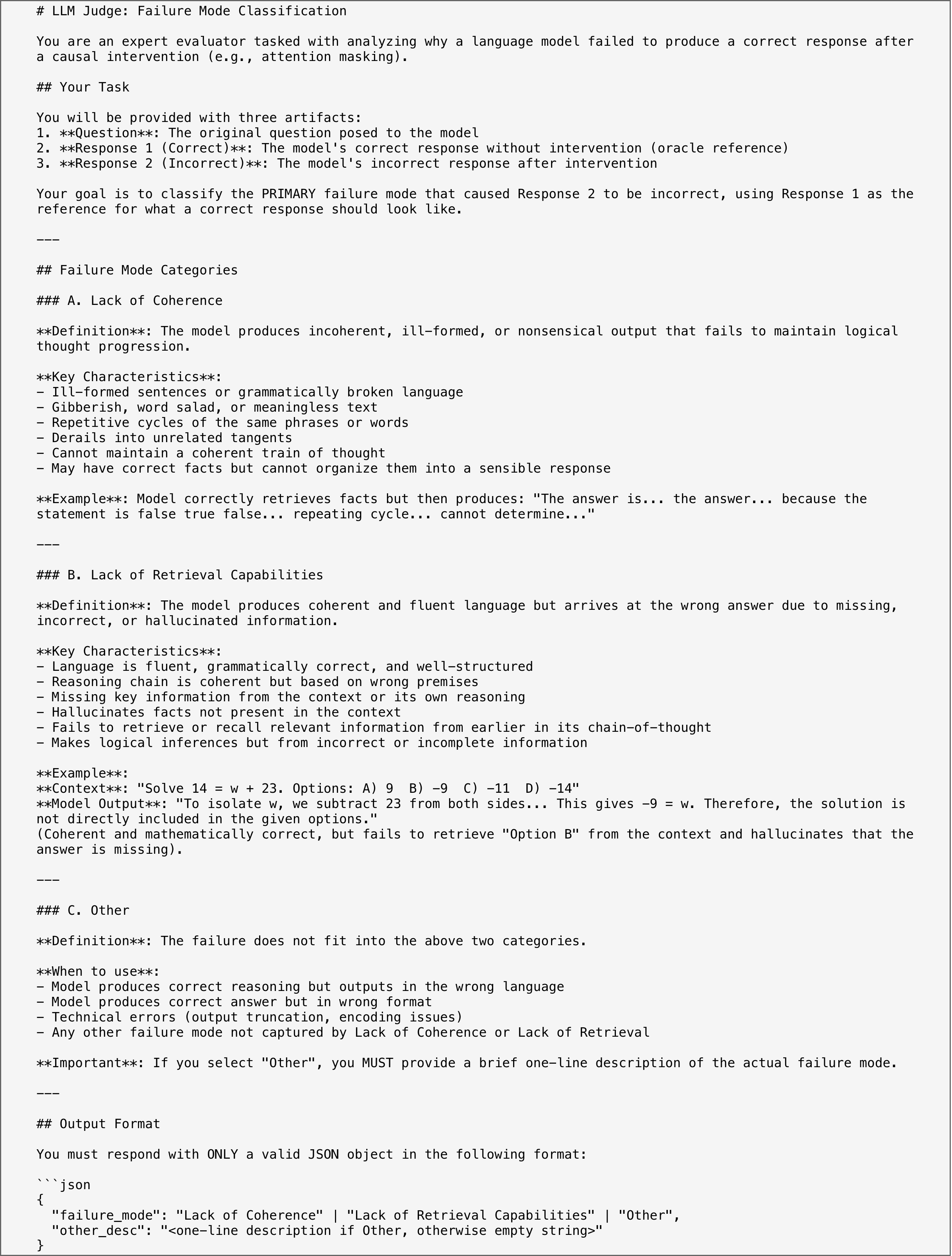}
    \caption{LLM-as-a-judge prompt used to categorize flipped samples into failure modes (Loss of Coherence, Failure to retrieve, Other). The judge is given the original question, the correct answer chain-of-thought from the No-Mask condition, and the model outputs under different masking settings, and is instructed to assign a single failure label along with a brief justification.}
    \label{fig:prompt_for_classifcation}
\end{figure}

\subsection{Human Validation of the LLM-as-a-judge}\label{appendix:qualitative_human_validation}
\camera{We use an LLM-as-a-judge to sort masking-induced answer flips into three failure modes: Loss of Coherence (LoC), Failure to Retrieve (RF), and Other. To establish that these labels are trustworthy, we validated the judge against a human annotator.}

\paragraph{Dataset.} \camera{We drew 50 sample from the zhen judge outputs Qwen2.5-7B, spanning both MMLU-ProX (38 samples) and MGSM (12 samples). The sample contain both RH and RTH masking results at k = 25. The sample contains 20 LoC, 24 RF, and 6 Other by the judge's labels.}

\paragraph{Annotation protocol.} \camera{The annotator was shown the same rubric and the same case material as the judge. The annotator had to assign one of the three failure modes. Annotation was blind.}

\paragraph{Evaluation and Results.} \camera{One of the authors annotated all 50 samples. The human annotator and the judge assigned the same failure mode on 39 of the 50 items (78.0\%). Correcting for chance agreement, this gives a Cohen's $\kappa$ of 0.633, which falls in the substantial band \cite{landis1977}. On the LoC-vs-RF distinction that Section~\ref{subsec:qualitative} relies on, the two agree on 37 of 40 items (92.5\%), with $\kappa = 0.851$. Table \ref{tab:human_judge_confusion} gives the full confusion matrix.}

\begin{table*}[h!]
  \centering
  \small
  \begin{tabular}{lcccc}
  \toprule
   & \multicolumn{3}{c}{\textbf{LLM Judge}} & \\
  \cmidrule(lr){2-4}
  \textbf{Human} & LoC & RF & Other & Total \\
  \midrule
  LoC   & \textbf{18} & 3           & 2          & 23 \\
  RF    & 0           & \textbf{19} & 2          & 21 \\
  Other & 2           & 2           & \textbf{2} & 6  \\
  \midrule
  Total & 20          & 24          & 6          & 50 \\
  \bottomrule
  \end{tabular}       
  \caption{Confusion matrix between the human annotator (rows) and the LLM judge (columns) on the 50 samples. Bold diagonal entries are agreements.}
  \label{tab:human_judge_confusion}
\end{table*}


\subsection{Qualitative Examples of failure modes on masking Retrieval-Transition heads}\label{appendix:sec_Qualitative}

\paragraph{Failure Mode 1: Lack of Coherence.} Figure \ref{fig:lack_of_coherence} shows the output comparison of Qwen2.5 7B for a German MMLU-ProX Math problem. The unmasked model outputs a concise and correct thought chain. However, when top-25 retrieval-transition heads for German ($RTH_{de}$) are masked, the model fails to reach the correct answer. This is even after the model retrieves the correct facts for both statements, correctly identifying the first as false and the second as true, but still unable to form a coherent thought. Instead, it devolves into repetitive, nonsensical cycles and fails to provide a final answer before hitting the decoding limit.

\begin{figure}[h!]
    \centering
    \includegraphics[width=0.9\linewidth]{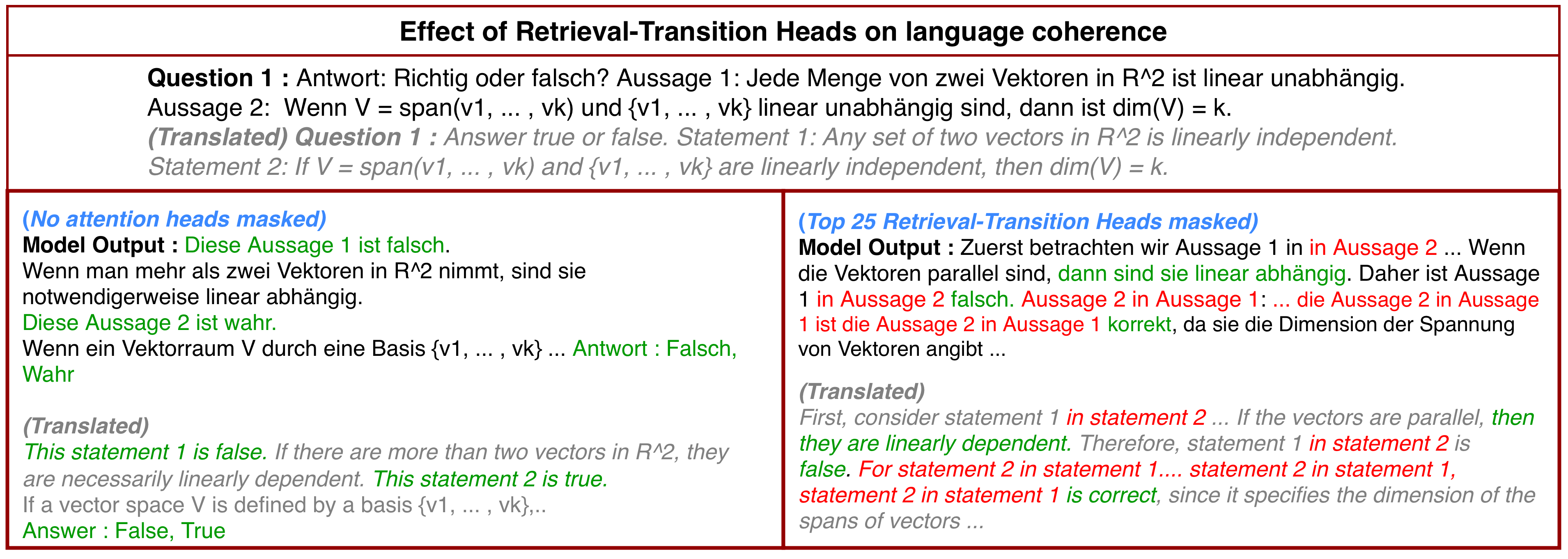}
    \caption{Masking RTHs disrupts the mapping from latent reasoning to target-language syntax. While the model may grasp the underlying fact, it fails to synthesize grammatical output, resulting in structural gibberish or character-level hallucinations.}
    \label{fig:lack_of_coherence}
\end{figure}

\paragraph{Failure Mode 2: Failure to Retrieve Capabilities.} Even for fairly simple problems, requiring little reasoning, masking RTH heads causes hallucinations and incomplete retrieval, like RH heads caused in long-context reasoning \cite{wu2024retrieval}. Figure \ref{fig:lack_of_retrieval} shows one such example using a simple Arithmetic in German. The model performs the arithmetic correctly and calculates the value of $-9$, but is unable to retrieve the corresponding option from the context due to top 25 most influential RTH heads being masked. It incorrectly concludes that the correct answer is not present in the options and ultimately starts hallucinating. Such failures are not observed when masking the top-25 RH for that language suggesting that RTH act as the essential bridge allowing an MLM to map internal reasoning back to specific target-language tokens and options.

\begin{figure}[h!]
    \centering
    \includegraphics[width=0.9\linewidth]{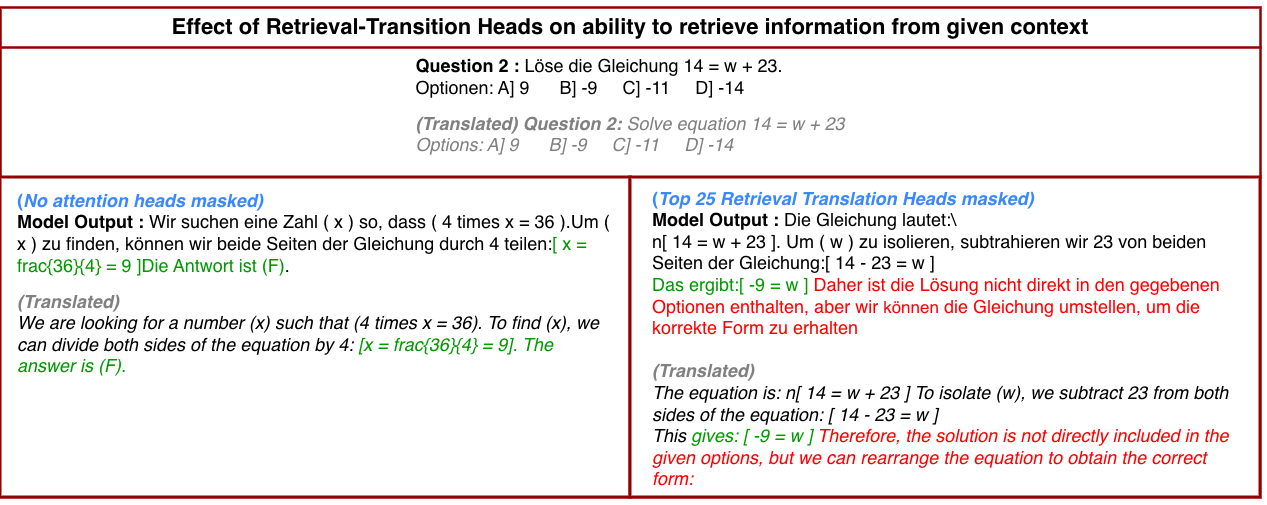}
    \caption{The model loses the ability to align internal reasoning with the prompt's constraints. Even if the logic remains correct, the model is unable look-back to the provided multiple-choice options, leading to answers that are ungrounded in the context.}
    \label{fig:lack_of_retrieval}
\end{figure}

\end{document}